\documentclass{article}

 \usepackage[preprint]{neurips_2026}

\usepackage[utf8]{inputenc} 
\usepackage[T1]{fontenc}    
\usepackage{hyperref}       
\usepackage{url}            
\usepackage{booktabs}       
\usepackage{amsfonts}       
\usepackage{nicefrac}       
\usepackage{microtype}      
\usepackage{xcolor}         

\usepackage{graphicx} 

\title{Towards Annotation-Free Validation of MLLMs: A Vision-Language Logical Consistency Metric}

\author{Ying Gu, Mei Chee Leong, Hui Li Tan, Shangbo Mao, Liyuan Li, Nancy Chen \\
Institute for Infocomm Research (I$^2$R),\\ 
    Agency for Science, Technology and Research (A*STAR),\\
    Singapore}

\begin{document}

\maketitle

\begin{abstract}
  Dominant accuracy evaluation might reward unwarranted guessing of Large Language Models~\cite{Kalai:nature2026}, and it might not be applicable to novel tasks for model validation without ground-truth (gt) annotation. Based on basic logic principle, we propose a novel framework to evaluate the vision-language logical consistency of MLLMs on both sufficient and necessary cause-effect relations. We define Vision-Language Logical Consistency Metric (VL-LCM) on traditional MC-VQA tests, and recent NaturalBench tests without the need for gt annotation. Through systematic experiments on representative VL benchmark MMMU and recent VL challenges like NaturalBench, we evaluated 11 recent open-source MLLMs from 4 frontier families. Our findings reveal that, despite significant progress of recent MLLMs on accuracy, logical consistency lags behind significantly. 
  Extensive evaluations on the correlations of VL-LCM with metrics on gt, the reliability of LCM, and the relation of VL-LCM with response distribution justify the validity and applicability of VL-LCM even without gt annotation. Our findings suggest that, beyond accuracy, logical consistency could be employed for both accuracy and reliability. VL-LCM can also be employed for MLLM selection, validation, and reliable answer justification in novel tasks without gt annotation.
\end{abstract}


\section{Introduction}

Recent progress in MLLM represents significant steps towards Artificial General Intelligence (AGI), with frequent release of updated frontier models and the rapid performance improvements on leaderboards and various vision-language (VL) benchmarks~\cite{zhao2025surveylargelanguagemodels, caffagni-etal-2024-revolution, fu2025mmecomprehensiveevaluationbenchmark, yue2024mmmu,li2024surveybenchmarksmultimodallarge}, as well as numerous applications~\cite{liu2024mmbenchmultimodalmodelallaround, wang2024comprehensivereviewmultimodallarge}. Despite their ground-breaking capability in both traditional vision tasks and recent complex multimodal problems~\cite{yue2024mmmu,ghosh2025exploringfrontiervisionlanguagemodels}, recent studies increasingly highlight the limitations of these models in terms of reliability and trustworthiness~\cite{10.3389/fnrgo.2023.1216440,zhang2025exploringgeneralizabilityfactualhallucination, he-etal-2025-evaluating}. 

Accuracy-based metrics are widely used to evaluate the progress of MLLMs in various VL tasks. Recent studies have revealed that accuracy-based evaluation might reward unwarranted guessing of LLM~\cite{Kalai:nature2026} and less effective to mirage reasoning where MLLM makes decision without vision input~\cite{asadi2026mirageillusionvisualunderstanding}. They may also not be applicable to novel tasks and applications in high-stakes domains as the ground-truth (gt) annotation is unavailable~\cite{Khan_2024_CVPR}.

\begin{figure}[t]
  \centering
\includegraphics[width=1.0\textwidth]{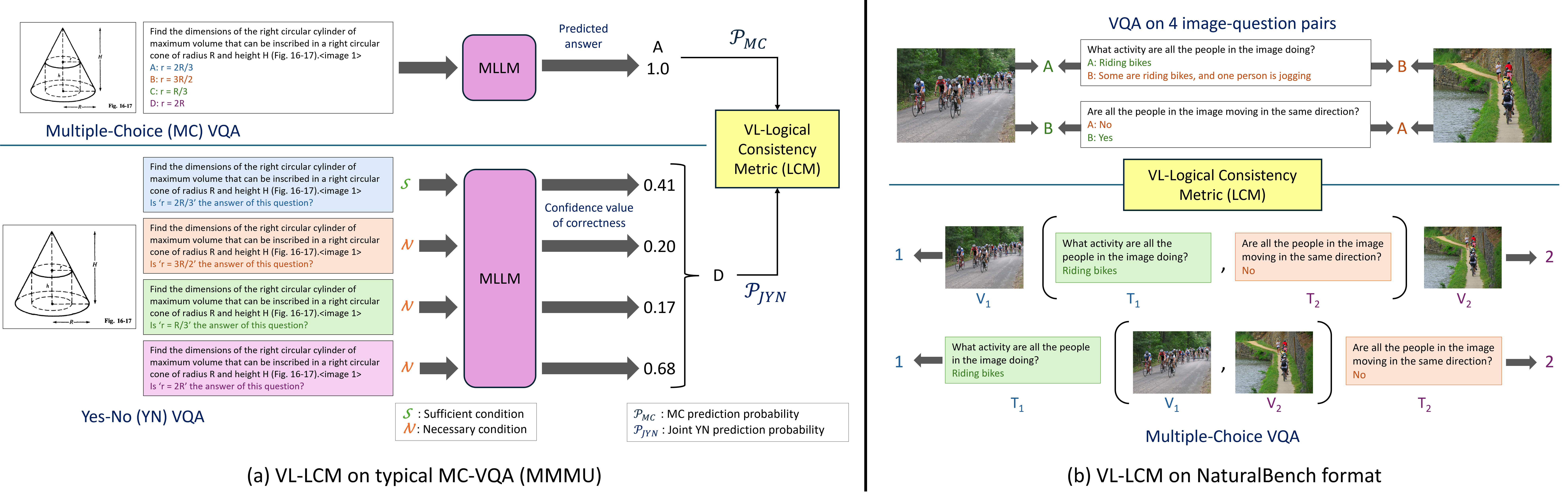} 
\caption{The illustration of our approach to compute VL-LCM on typical MC format (eg., MMMU) and recent NaturalBench format. In the left side, the upper part shows the normal MC-VQA test and the prediction with probability. In the lower part, the MC problem is separated as a group of Yes/No (YN) tests, and the joint probabilities of each choice are obtained on the MLLM's predictions. Finally, VL-LCM is computed on the MC and YN tests. In the right side, the upper part shows the YN tests on four image-question pair as defined in NaturalBench. The lower part illustrates the MC tests derived from NaturalBench format, where each image with two text expressions (upper) and each text expression with two images are fed to MLLM to select the correct one. VL-LCM is computed on the consistency of the YN and MC tests.}
\label{fig:framework}
\end{figure}

Based on basic logic principle, we propose a novel framework to evaluate the vision-language logical consistency on both sufficient and necessary cause-effect relations, and the formulation of Vision-Language Logic Consistency Metric (VL-LCM) on typical MC-VQA and recent NaturalBench formats, as illustrated in Figure~\ref{fig:framework}. Instead of evaluating the self-consistency on a group of tests with sufficient conditions ({\em i.e.} $p \to q$), {\em e.g.} in~\cite{NEURIPS2024_d6f094ba, Khan_2024_CVPR, chou-etal-2025-mm}, we propose to test the MLLM on both logical sufficiency and necessity, and compute the logic consistency on both MC-VQA and YN-VQA (yes-no VQA). 
Based on the logic formulation of cause-effect relationship, we formulate the statistical prediction probabilities by MLLM as logic predictions of the answers on both sufficient and necessary conditions on the selective VQA tests. Then, on the typical MC-VQA and recent NaturalBench formats, we formulate the vision-language logic consistency as logic inference on the combination of the MC and YN tests on both logical sufficiency and necessity. The proposed VL-LCM can be obtained without gt annotation, applicable for annotation-free model ranking, selection and validation in novel tasks in real-world applications. As the VL-LCM is defined as per-sample measurement, it could also be used in online validation.

Systematic experiments and extensive evaluations are performed on representative VL benchmark MMMU~\cite{yue2024mmmu}, and recent VL challenges like NegBench~\cite{alhamoud2025vision},  ConBench~\cite{zhang2024conbench}, NaturalBench~\cite{li2024naturalbench}, as well as a new NatConBench automatically generated from ConBench on NaturalBench format, with 11 recent open-source MLLMs from four frontier families. Our experimental results reveal that: (a) VL-LCM might provide deep insights on both accuracy and reliability beyond accuracy; (b) VL-LCM is strongly correlated to gt-based F1 on MC and YN tests, validating the effectiveness for annotation-free validation; (c) VL-LCM could be used for reliable answer selection and justification.

The main contributions of this paper can be summarized as: (1) a general framework for logic consistency evaluation of MLLM on both sufficient and necessary conditions, and VL-LCM on typical MC-VQA and recent NaturalBench formats; (2) systematic experiments on four representative and recent challenging VL benchmarks, with 11 recent MLLMs of 4 frontier families; (3) new findings on the effectiveness of VL-LCM for annotation-free validation of MLLM.

\section{Related Work}
\noindent {\bf Annotation-free validation of Large Models}: 
Since Large Language Models (LLMs) are trained with RLHF \cite{ouyang2022_RLHF}, they often show strong human alignment, making them a practical judge for scalable evaluation. \cite{zheng2023_llm-as-a-judge} systematically studies the use of strong LLMs such as GPT-4 to evaluate other LLMs. LLM-based evaluators are now widely used in natural language generation tasks \cite{wang2023_chatgpt_nlg_evaluator}\cite{liu2023_G-eval}\cite{chiang2023_automatic_LLM}\cite{verga2024_judges_with_juries}.
Inspired by this trend in LLM evaluation, LLM-as-judge has also been applied to open-ended MLLM evaluation \cite{yu2024_MM-Vet}. 
However, incorporating external judge models may introduce systematic biases \cite{wang2024_LLM_fair}, such as sensitivity to response ordering. This limitation underscores the need for more reliable annotation-free methods for evaluating MLLMs.

\noindent {\bf Self-consistency on MLLM}: Self-consistency of MLLM on VQA tests has emerged as a crucial issue in recent studies~\cite{Tascon-Morales_2023_CVPR}. The representative approach is to build a new benchmark dataset with clusters of related visual questions, and evaluate the consistency of the answers~\cite{jimenez-etal-2022-carets}. Recently, in~\cite{NEURIPS2024_d6f094ba}, a ConBench dataset is established, where, on each problem, there are three related VQA tests on the formats of yes/no, MC, and open-end reply. A new metric ConScore[D] is defined which requires that all three answers are correct. In NaturalBench~\cite{li2024naturalbench}, a pair of images and two questions with alternating answers on the two images are grouped, where the images and the questions are visually and semantically similar. The joint accuracy on the four image-question pairs is counted for performance evaluation. Approaches are proposed which, on a given visual question, automatically generating a cluster of neighborhood questions on linguistic variations~\cite{Khan_2024_CVPR,tasconmorales2023logicalimplicationsvisualquestion}, or generate associated caption-image pairs iteratively~\cite{cao-etal-2024-introducing}, and the self-consistency of the answers is measured. Existing methods require building a new dataset with gt annotation, and the self-consistency is evaluated on logical sufficiency and accuracy metrics~\cite{jimenez-etal-2022-carets}.

\noindent {\bf Logic consistency on LLMs}: Logic consistency of LLM requires the model's responses to be coherent, factually correct, and logically sound~\cite{Mitchell2022}, as LLMs are prone to generating responses contradicting themselves across different questions of a related problem~\cite{creswell2022selectioninferenceexploitinglargelanguage, 10.24963/ijcai.2025/1155, ghosh2025logicalconsistencylargelanguage}. Many methods have been proposed to improve the logic consistency of LLMs, including solver-based, prompt-based, and fine-tuning methods to address various types of logic violations such as negation, implication, transitivity, factuality and composites~\cite{10.24963/ijcai.2025/1155, calanzone2024logicallyconsistentlanguagemodels}. A widely adopted framework for consistency of LLM is defined on a collection of examples with statistics of global violation and conditional violation~\cite{li-etal-2019-logic}. Existing efforts still focus on semantic consistency of language expressions.

\section{Methodology}

\subsection{Problem Statement}

The Multimodal LLM is an extension of LLM, which employs a pre-trained LLM as backbone. The general architecture consists of image and text encoders and vision-text embeddings~\cite{Li_2025_CVPR}. The visual tokens are inserted into text sequence and fed to LLM to predict next tokens auto-regressively. Let $t_1^v, \cdots, t_n^v$ be the sequence of extracted and embedded visual tokens from image $V$, and $t_1^t, \cdots, t_k^t$ be the fed text tokens of language input $T$. The MLLM generates next text token based on learned co-occurrence patterns in previous input data. The probability of the statistical prediction can be expressed as
\begin{equation}\label{eq:spp}
p(t_{k+1} | V, T) = p(t_{k+1}^t|t_1^v, \cdots, t_n^v, t_1^t, \cdots, t_k^t). 
\end{equation}

The statistical prediction probability of a MLLM is established by learning the syntax, semantic, and world knowledge through vision-language tasks for training. The dominant vision-language tasks are various Visual Question Answering (VQA) tasks~\cite{Li_2025_CVPR,fu2024mmesurveycomprehensivesurveyevaluation}. In general, the VQA task is a logic inference task: $V,T \to a$, {\em i.e.}, given an image $V$ and a text query $T$, predicting a correct answer $a$. Prediction probability is well calibrated with accuracy frequency for LLMs on MC and YN (or True/False) tasks~\cite{kadavath2022languagemodelsmostlyknow}.

In logic theory of cause-effect relation~\cite{Gomes:24}, such VQA training sample represents the sufficient condition of the logic inference, {\em i.e.}, the input $V$ and $T$ is a sufficient cause of the answer $a$, but may not be necessary. On the basic logic principle, if a MLLM is reliable on a specific problem of vision-language understanding, its behavior should be consistent on both sufficient and necessary conditions. We propose a general framework of logic consistency to evaluate the MLLM on both sufficient and necessary cause-effect relations even without gt annotations. On a VQA test $V,T \to a$, we jointly test the model with queries $\neg V, T \to \neg a$ and $V, \neg T \to \neg a$, where the former one tests the model on sufficient condition, and the latter two test the model on necessary condition. On logic cause-effect formulation, the tests can be expressed as:\\ 
\noindent $\bullet$ $V,T \to a$: {\em If image $V$ and query $T$ then answer $a$};\\
\noindent $\bullet$ $\neg V,T \to \neg a$: {\em If not image $V$ ($\neg V$) but query $T$ then not answer $a$ ($\neg a$)};\\
\noindent $\bullet$ $V,\neg T \to \neg a$: {\em If image $V$ but not query $T$ ($\neg T$) then not answer $a$ ($\neg a$)}.\\
The statistical prediction probabilities on these tests can be expressed as $P(a|V,T)$, $P(\neg a| \neg V, T)$ and $P(\neg a | V, \neg T)$. The probability on a MC-VQA sample of $K$ choices can also be expressed as $P(a|V,T_1, \cdots, T_K)$. Under this framework, we propose a metric, {\em i.e.}, VL-LCM, to measure the logic consistency of the MLLM's performance on the selective VQA tests even without gt annotation.


\subsection{Vision-Language Logical Consistency Metric on MC-VQA}\label{sec:lcm-mcvqa}

A typical MC-VQA sample consists of a query image ($V$), a question ($T$), and a few potential answers ($A_k$), where $k \in [1, K]$ and in most cases $K$=4. The MLLM is prompted to predict the correct answer ($a^{\ast}$). Since all potential answers are provided in sequence when prompted to select the correct answer, the probability of MC prediction for each potential choice can be expressed as
\begin{equation}
P_{MC}(a_k) = P_m(a_k|V,T,A_1, \cdots, A_K),
\end{equation} 
where $P_m$ denotes the statistical prediction probability of the model.

On MC-VQA format, as the MLLM is prompted to select one from $K$ choices, it might exploit shortcut cues beyond the vision-language patterns. If we test MLLM with the question and each choice separately by asking if the choice is right or not, we may obtain the MLLM's response merely on its learned vision-language knowledge. When tested on choice $A_k$ separately (yes-no (YN) of the $k$-choice), the prediction probability of the confirmed answer and its negation can be expressed as
\begin{equation}
\left\{
  \begin{array}{ll}
    P_{YN}(a_k) & =  P_m(a_k|V,T,A_k),\\
    P_{YN}(\neg a_k) & = 1 -  P_m(a_k|V,T,A_k).
  \end{array}
\right.
\end{equation}
$P_m(a_k|V,T,A_k)$ is generated purely on learned vision-language patterns. On the set of $K$ YN tests, if the right choice is $A_k$, the set of YN tests can be expressed logically as\\
\noindent $\bullet$ $V,T, A_k \to a_k$: {\em If image $V$, question $T$ and choice $A_k$ then answer $a_k$};\\
\noindent $\bullet$ $V,T, A_i \to \neg a_k$ ($i \ne k$) or $V,T, \neg A_k \to \neg a_k$: {\em If image $V$, question $T$, but not choice $A_k$ then not answer $a_k$ ($\neg a_k$)};\\
where the former is a test on sufficient condition and the latter is a test on necessary condition, and $a_k$ indicates yes($A_k$). In each MC-VQA test, there is only one right answer. Hence, the logic probabilities on sufficient and necessary conditions are computed as
\begin{equation}
\left\{
  \begin{array}{ll}
    P_{suf}(a_k) & =  P_{YN}(a_k|V,T,A_k),\\
    P_{nec}(\neg a_k) & = \min_{i \in [1,K], i \neq k} \left[1 - P_{YN}(a_i|V,T,A_i) \right]
  \end{array}
\right.    
\end{equation}
The joint probability for choice $A_k$ can be computed as
\begin{equation}\label{eq:joint_sc}
P_{JYN}(a_k) = \left( P_{suf}(a_k) P_{nec}(\neg a_k) \right)^{1/2} 
\end{equation}
where the geometric mean is used for normalization as done in~\cite{malinin2020uncertainty}. Eq.~(\ref{eq:joint_sc}) means that the correct answer is $a_k$ and not others, which measures the logical consistency on both sufficient and necessary conditions when tested on each choice separately.

If a MLLM has learned the visual-language knowledge on a MC-VQA problem correctly and reliably, its performance on MC test and the set of YN tests should be logically consistent. Hence, since there is only one correct answer in the $K$ choices, we can define a visual-language logical consistency metric (VL-LCM) as
\begin{equation}\label{eq:vllcm_mc}
P_{LC} = \max_{k \in [1,K]}  \left\{ (P_{MC}(a_k) P_{JYN}(a_k))^{1/2} \right\}.
\end{equation}
If a MLLM is able to predict the correct answer on both MC test and joint YN tests, the $P_{LC}$ is high and close to 1.0, otherwise, the $P_{LC}$ is low and close to 0.0. $P_{LC}$ is obtained for each MC-VQA test without gt annotation. Hence, $P_{LC}$ could be used to validate the accuracy and reliability of the MLLM performance even without gt annotation. If the gt choice is known, we can obtain $P_{LC}^{\ast}$ with $a_k=a^{\ast}$ as the LCM on gt. Naturally, $P_{LC}^{\ast} \leq P_{LC}$, and in ablation study, it is found that the larger the $P_{LC}$ score, the closer of $P_{LC}^{\ast}$ to $P_{LC}$.

\subsection{Vision-Language Logical Consistency Metric on NaturalBench format}\label{sec:lcm-nb}
In MC-VQA format, only one choice is the correct answer. The imbalanced truth and negation answers might lead to the biased evaluation of MLLM. In recent NaturalBench~\cite{li2024naturalbench}, a balanced MC-VQA format is proposed. Each testing unit consists of two images and two questions with alternating answers, where the images and questions are visually and semantically similar. Each image-question pair can be considered as a test of logical sufficiency ({\em i.e.} $V,T \to a$).

Under the framework of logic consistency on both sufficient and necessary conditions, we propose to perform both YN and MC tests on one unit sample of NaturalBench, and evaluate the logic consistency between YN and MC tests without gt annotation. Formally, in one test sample, there are two images $V_1$ and $V_2$, which are visually similar, and two questions $T_1$ and $T_2$, which are semantically similar, where for each question, the answer ($a_1$) is `yes' or `A' on one image, and ($a_2$) `no' or `B' on another image. As the two images and two questions are selected with alternating answers, on logic formulation, the YN tests on $a_1$ can be expressed as\\
\noindent $\bullet$ $V,T \to a$ ({\em e.g.}, $V_1,T_1 \to a_1$ or $V_2,T_2 \to a_1$): {\em If image $V$ and question $T$ then answer $a$};\\
\noindent $\bullet$ $V,\neg T \to \neg a$ ({\em e.g.}, $V_1, T_2 \to \neg a_1$): {\em If image $V$ but not question $T$ then not answer $a$ ($\neg a$)};\\
\noindent $\bullet$ $\neg V, T \to \neg a$ ({\em e.g.}, $V_2, T_1 \to \neg a_1$): {\em If question $T$ but not image $V$ then not answer $a$ ($\neg a$)};\\
where the first is a test on sufficient condition and the remaining two are tests on necessary conditions. Same logic conditions are applied to $a_2$. Then, on the yes-no (YN) VQA tests, the statistical prediction probability of the right answer and its negations are expressed as
\begin{equation}\label{eq:naturalbench_sc}
    P_{YN}(a) =  P_m(a|V,T),\quad
    P_{YN}(\neg a) = 1 -  P_m(a|V,\neg T) ~\textrm{or}~
    P_{YN}(\neg a) = 1 -  P_m(a|\neg V,T).
\end{equation}

Again, as the two images and two questions are selected with alternating answers, we can also design a set of MC-VQA tests on each sample, {\em i.e.}, on one image and two questions, choose the correct one, and on one question and two images, choose the correct one, as illustrated in Eq.~(\ref{fig:vl-lcm}). The tests are\\
\noindent (a) $V_1,~T_1 ~or~ T_2$: On image $V_1$, choose text statement $T_1$ or $T_2$;\\
\noindent (b) $V_2,~T_1 ~or~ T_2$: On image $V_2$, choose text statement $T_1$ or $T_2$;\\
\noindent (c) $T_1,~V_1 ~or~ V_2$: On text statement $T_1$, choose image $V_1$ or $V_2$;\\
\noindent (d) $T_2,~V_1 ~or~ V_2$: On text statement $T_2$, choose image $V_1$ or $V_2$.\\
Let us focus on MC test (a) first. When a MLLM is tested on problem (a), it predicts the choice of the right statement with the statistical prediction probabilities
\begin{equation}
\left\{
  \begin{array}{ll}
    P_{MC}^{(a)}(c_1) & = P(c_1)P(\neg c_2) =  P_m(c_1|V_1,T_1,T_2) (1-P_m(c_2|V_1,T_1,T_2))\\
    P_{MC}^{(a)}(c_2) & = P(c_2) P(\neg c_1) = P_m(c_2|V_1,T_1,T_2) (1-P_m(c_1|V_1,T_1,T_2))
  \end{array}
\right.
\end{equation}
where $c_1$ indicates that the pairs $V_1$-$T_1$ and $V_2$-$T_2$ are correct pairs, and $c_2$ means that the pairs $V_1$-$T_2$ and $V_2$-$T_1$ are correct pairs. On the other side, from the YN tests on the sample, we can compute the joint probabilities for choices $c_1$ and $c_2$ on MC test (a) as
\begin{equation}\label{eq:jsc_acc_naturalbench}
\left\{
  \begin{array}{ll}
    P_{JYN}^{(a)}(c_1) & = P(c_1|V,T) P(\neg c_1 | V, \neg T) = P_m(c_1|V_1,T_1) (1-P_m(c_1|V_1,T_2))\\
    P_{JYN}^{(a)}(c_2) & = P(c_2|V,T) P(\neg c_2|V,\neg T) = P_m(c_2|V_1,T_2) (1-P_m(c_2|V_1,T_1)).
  \end{array}
\right.
\end{equation}

On the MC and YN tests on test (a), we can define a visual-language logical consistency metric (VL-LCM) as
\begin{equation}\label{eq:nba-lcm}
P_{LC}^{(a)} = \max_{l \in [1,2]} \left\{ [P_{MC}^{(a)}(c_l)  P_{JYN}^{(a)}(c_l)]^{1/4} \right\},
\end{equation}
as only one choice is right, $P_{LC}^{(a)} \in [0,1]$. Similar, we can obtain the VL-LCMs on tests (b), (c) and (d). The final VL-LCM on a sample unit of NaturalBench format is obtained as
\begin{equation}\label{eq:lcm_naturalbench}
P_{LC} = \frac{1}{4} \left( P_{LC}^{(a)} + P_{LC}^{(b)} + P_{LC}^{(c)} + P_{LC}^{(d)} \right).
\end{equation}

Again, if the gt choice $c^{\ast}$ is known ($c_1$ or $c_2$), we can obtain $P_{LC}^{(a)\ast}$ and then $P_{LC}^{\ast}$ as LCM on gt.

In real-world application for a new vision-language understanding task, we may just have a set of separated image-question pairs without gt answers. On NaturalBench format, we can randomly select two image-question pairs with less chance of being true logically for crossing image-question pairs, which forms a testing unit. A set of such testing units can be employed for annotation-free validation of MLLM with LCM under both logical sufficient and necessary conditions.

\section{Experiments}

\noindent {\bf MLLMs}: The models employed in this study are: InternVL-2.0, 2.5, 3.0, and 3.5 8B models~\cite{team2024internvl2,chen2024internvl-25,zhu2025internvl-3,wang2025internvl-35}; Qwen-VL-7B-Chat, Qwen 2.0, 2.5 VL-7B-Instruct and 3.0 8B-Instruct models~\cite{bai2023qwen-vl,wang2024qwen2-vl,bai2025qwen25-vl,yang2025qwen3-vl}, LLaVA-1.5-13B, 1.6-13B model~\cite{liu2024llava15}, and Gemma-3.0-12B~\cite{gemma_2025}. In this study, we focus on open-source frontier MLLMs of moderate size (7-8B) for availability and computational efficiency in real-world applications.

\noindent {\bf Metrics}:
The metrics used for the experiments are: {\bf Acc}: accuracy obtained on the protocol of the corresponding benchmark, and {\bf J-Acc}: accuracy obtained on Eq.~(\ref{eq:joint_sc}) or (\ref{eq:jsc_acc_naturalbench}). We also propose a {\bf F1} metric of Acc and J-Acc, which balances the accuracy on MC and consistency on joint YN tests. They are obtained on gt annotations. Annotation-free {\bf VL-LCM} is obtained on Eq.~(\ref{eq:vllcm_mc}) or (\ref{eq:lcm_naturalbench}). The correlations of VL-LCM with the three metrics on gt annotation are evaluated on three representative correlation coefficients.

\noindent {\bf Prompting strategy}:
For each benchmark, we use the official prompt format for the main evaluation and an adapted prompt format for additional evaluation. Specifically, NegBench, ConBench, and MMMU use official prompts for MC testing and adapted prompts for YN testing, while NaturalBench uses official prompts for YN testing and adapted prompts for MC testing. Recent studies of prompt skills are considered for stable performance~\cite{mohanty2025futuremllmpromptingadaptive}. Details are provided in Appendix.

\subsection{Benchmarks}
Experiments are conducted on the representative VL benchmark MMMU~\cite{yue2024mmmu}, and three recent VL challenges from recent top AI conferences, {\em i.e.}, NegBench~\cite{alhamoud2025vision}, ConBench~\cite{zhang2024conbench} and NaturalBench~\cite{li2024naturalbench}, as well as new benchmark NatConBench automatically generated from ConBench on NaturalBench format. 
Details of public benchmarks and NatConBench construction are presented in Appendix. 
 
\noindent {\bf NatConBench}: Leveraging the diverse capability assessment of ConBench and the effective evaluation pipeline of NaturalBench, we introduce a novel dataset, NatConBench. We adapt NaturalBench protocol to ConBench, ensuring consistent evaluation within diverse categories while enforcing reliance on visual input. We generated 50 QA pairs for each of the 19 categories in ConBench and constructed a total of 1,850 samples, with balanced split between True/False and multiple-choice formats to form NatConBench.

\subsection{Experimental results and observations}

\begin{table}
  \caption{Experimental results on NegBench benchmark.}
  \label{tab:negbench}
   \footnotesize
  \centering
  \begin{tabular}{l|llll|llll}
    \toprule
     & \multicolumn{4}{c|}{COCO} & \multicolumn{4}{c}{VOC2007}  \\
    \cmidrule(r){2-9}
    MLLM & Acc/R & J-Acc & F1/R & LCM/R & Acc/R & J-Acc & F1/R & LCM/R \\
    \midrule
    Gemma3.0-12B & 71.98/6 & 42.07/8 & .5310/7 & .4579/9 & 87.50/5 & 60.09/8 & .7125/6 & .6462/6 \\
    InternVL-2.0-8B & 48.78/9 & 45.59/6 & .4713/8 & .2683/9 & 58.68/10 & 70.11/6 & .6388/9 & .4432/9 \\
    InternVL-2.5-8B & 91.21/2 & 48.28/4 & .6313/4 & .4654/4 & 92.01/3 & 74.18/2 & .8214/1 & .6812/3 \\
    InternVL-3.0-8B & 93.19/1 & 54.13/1 & .6848/1 & .5642/1 & 95.37/1 & 70.34/5 & .8097/3 & .6914/2 \\
    InternVL-3.5-8B & 84.29/3 & 53.52/2 & .6547/2 & .4899/3 & 92.21/2 & 72.51/3 & .8118/2 & .6759/4 \\
    LLaVA-1.5-13B & 62.06/8 & 46.75/5 & .5333/6 & .4009/7 & 77.38/8 & 74.78/1 & .7606/5 & .6553/5 \\
    LLaVA-1.6-13B & 38.98/10 & 37.88/9 & .3842/9 & .3061/8 & 65.28/9 & 58.68/10 & .6180/10 & .4417/10\\
    Qwen-VL-7B-Chat & 24.26/11 & 36.91/10 & .2928/11 & .1426/11 & 23.71/11 & 61.58/7 & .3424/11 & .1900/11 \\
    Qwen2.0-VL-7B & 69.68/7 & 20.87/11 & .3211/10 & .1870/10 & 85.35/6 & 54.82/11 & .6676/8 & .5327/8 \\
    Qwen2.5-VL-7B & 81.65/4 & 51.50/3 & .6317/3 & .5235/2 & 91.97/4 & 56.99/10 & .7037/7 & .5717/7 \\
    Qwen3.0-VL-8B & 77.53/5 & 45.45/7 & .5731/5 & .4570/6 & 84.08/7 & 70.40/4 & .7664/4 & .7232/1 \\
    \midrule    
    Pearson's $r$ & .8265 & .8079 & .9615 & & .9194 & .4718 & .9671 &  \\
    Spearman's $\rho$ & .8909 & .8727 & .9545 &  & .7000 & .6182 & .9182 &  \\
    Kendall's $\tau$ & .7091 & .7091 & .8545 & & .5636 & .3818 & .8182 &  \\
    \bottomrule
  \end{tabular}
\end{table}
\noindent {\bf Results on NegBench}: The experimental results on NegBench are presented in Table~\ref{tab:negbench}, where `Acc' column shows the percentages of accuracy on MC test, and `LCM' column shows the VL-LCM scores. `/R' means the ranking on the corresponding metric. 
In Table~\ref{tab:negbench}, one can observe large progress of recent MLLMs. On COCO, LLaVA-1.5 gets 62.06\%, Gemma3.0 reaches 72.0\%, Qwen2.5-VL obtains 81.65\%, and InternVL-3.0 has achieved 93.19\% of accuracy rate, a great jump from 54\%~\cite{alhamoud2025vision}. On VOC2007, the SOTA accuracy rate has been increased to 95.37\% by InternVL-3.0. However, one may observe that the corresponding LCM scores are still much lower, mostly around 20\% to 40\% behind the corresponding accuracy rate. As examples, with InternVL-3.0, the Acc reaches 0.9319 on COCO, but LCM score is 0.5642, and Acc score is 0.9537 but LCM score is 0.6914 on VOC2007. To justify the observations on LCM, we introduce the joint accuracy rate (J-Acc) on the YN tests with $P_{JYN}(gt) > 0.5$ on Eq.~(\ref{eq:joint_sc}). The accurate answer on joint YN tests should be obtained if the model predicts the correct answer on gt choice and rejects all other choices when tested one-by-one independently. In Table~\ref{tab:negbench}, one can observe the large drop of J-Acc scores wrt to the corresponding Acc scores when evaluated on both logical sufficient and necessary conditions, justifying the low scores of LCM. To balance the MC and JYN tests, we propose a F1 metric of MC-Acc and JYN-Acc. In the table, it can be found that the LCM scores are much closer to the F1 scores than the Acc scores for the corresponding models. F1 is obtained on the statistics of MC-Acc and JYN-Acc on the full evaluation set, where MC-Acc and JYN-Acc are accumulated independently. VL-LCM is computed on per-sample jointly on MC and JYN performance on Eq.~(\ref{eq:vllcm_mc}) and then summed up over the full set. Hence, LCM score is more accurate and lower than F1 score. In the bottom three rows of the table, one can find the very strong correlations between F1 and LCM on both measurement by Pearson's $r$ and ranking by Spearman's $\rho$ and Kendall's $\tau$, justifying the validity of VL-LCM even without gt annotation.

\begin{table}
  \caption{Experimental results on ConBench and MMMU benchmarks.}
  \label{tab:ConBench-MMMU}
  \footnotesize
  \centering
  \begin{tabular}{l|llll|llll}
    \toprule
     & \multicolumn{4}{c|}{ConBench} & \multicolumn{4}{c}{MMMU}  \\
    \cmidrule(r){2-9}
    MLLM & Acc/R & J-Acc & F1/R & LCM/R & Acc/R & J-Acc & F1/R & LCM/R \\
    \midrule
    Gemma3.0-12B & 70.76/7 & 46.18/1 & .5589/1 & .5391/1 & 51.36/6 & 10.63/9 & .1761/9 & .1216/9 \\
    InternVL-2.0-8B & 64.52/8 & 28.05/5 & .3910/4 & .3082/6 & 45.45/8 & 15.70/5 & .2334/4 & .1634/6 \\
    InternVL-2.5-8B & 72.05/4 & 21.80/7 & .3348/7 & .2786/8 & 50.65/7 & 17.95/3 & .2650/3 & .1790/5 \\
    InternVL-3.0-8B & 71.46/6 & 44.20/2 & .5462/2 & .5101/2 & 55.37/2 & 12.51/7 & .2042/7 & .1279/7 \\
    InternVL-3.5-8B & 74.53/2 & 18.83/9 & .3006/9 & .3124/4 & 57.73/1 & 26.45/1 & .3628/1 & .3404/2 \\
    LLaVA-1.5-13B & 53.02/10 & 13.38/11 & .2137/11 & .1312/11 & 38.25/9 & 08.38/10 & .1375/10 & .0851/10 \\
    LLaVA-1.6-13B & 46.78/11 & 27.25/6 & .3444/6 & .2233/10 & 37.07/11 & 16.06/4 & .2242/6 & .2077/3 \\
    Qwen-VL-7B-Chat & 56.89/9 & 28.84/4 & .3828/5 & .2807/7 & 37.90/10 & 03.54/11 & .0648/11 & .0281/11 \\
    Qwen2.0-VL-7B & 71.75/5 & 16.55/10 & .2690/10 & .2337/9 & 52.54/5 & 11.45/8 & .1881/8 & .1263/8 \\
    Qwen2.5-VL-7B & 73.54/3 & 39.15/3 & .5110/3 & .4703/3 & 54.07/4 & 14.52/6 & .2290/5 & .1795/4 \\
    Qwen3.0-VL-8B & 76.51/1 & 21.41/8 & .3345/8 & .3117/5 & 54.90/3 & 22.90/2 & .3232/2 & .4013/1 \\
    \midrule    
    Pearson's $r$ & .5514 & .9156 & .9521 & & .5382 & .9388 & .9249 & \\
    Spearman's $\rho$ & .4818 & .7182 & .7273 & & .4455 & .9455 & .9091 & \\
    Kendall's $\tau$ & .3091 & .5636 & .6000 & & .3818 & .8545 & .7818 & \\
    \bottomrule
  \end{tabular}
\end{table}
\noindent {\bf Results on ConBench and MMMU}: COCO and VOC2007 are two traditional vision datasets with long histories, and also widely used to build various vision-language benchmarks. Hence, pre-trained on vast public data sources, the recent MLLM is able to achieve very high accuracy on NegBench. Beyond the task of vision-language perception in NegBench, ConBench and MMMU have extended to comprehensive tasks of vision-language reasoning and knowledge with much larger coverage of image types. Therefore, they are more challenging and difficult than NegBench. The experimental results on ConBench and MMMU are presented in Table~\ref{tab:ConBench-MMMU}. One can observe the lower accuracy rates compared with those on NegBench. Nonetheless, compared with the SOTA performance when the Benchmarks were released, great progress have been achieved by frontier MLLMs, especially on MMMU. On ConBench, the recent models from InternVL family have improved the Acc from 64.5\% to 74.5\%, and ones from Qwen family have increased the Acc from 56.9\% to 76.5\%. On MMMU, the models from InternVL family have improved the Acc from 45.5\% to 57.7\%, and ones from Qwen family have increased the Acc from 37.9\% to 54.9\%. The improvements on Acc are continuous and consistent as accuracy is one dominant metric for leaderboard. Similar to the observations on NegBench, one can find that the LCM scores are still very low compared with Acc, and often fluctuated with later version released. Again, this observation is justified by J-Acc scores. Due to the poor and unstable performance of logic consistency on logical sufficient and necessary conditions, LCM is weakly correlated with Acc but strongly correlated with J-Acc and F1, especially with F1, as indicated in the three bottom rows in Table~\ref{tab:ConBench-MMMU}. This finding validates the effectiveness of LCM as an annotation-free metric for both accuracy and reliability. 

\begin{table}
  \caption{Experimental results on NaturalBench and NatConBench benchmarks.}
  \label{tab:NaturalBench-NatConBench}
  \footnotesize
  \centering
  \begin{tabular}{l|llll|llll}
    \toprule
     & \multicolumn{4}{c|}{NaturalBench} & \multicolumn{4}{c}{NatConBench}  \\
    \cmidrule(r){2-9}
    MLLM & Acc/R & J-Acc & F1/R & LCM/R & Acc/R & J-Acc & F1/R & LCM/R \\
    \midrule
    Gemma3.0-12B & 74.84/5 & 54.53/5 & .6309/5 & .4179/6 & 80.98/4 & 65.92/3 & .7268/3 & .6011/1 \\
    InternVL-2.0-8B & 74.22/6 & 53.87/6 & .6243/6 & .4074/7 & 70.70/8 & 44.30/10 & .5447/10 & .3776/8 \\
    InternVL-2.5-8B & 78.70/3 & 61.08/3 & .6878/3 & .4845/1 & 72.23/7 & 48.16/7 & .5779/7 & .4290/7 \\
    InternVL-3.0-8B & 79.80/1 & 62.92/1 & .7036/1 & .4825/2 & 73.85/5 & 54.41/5 & .6266/5 & .4918/4 \\
    InternVL-3.5-8B & 71.41/8 & 49.71/8 & .5862/8 & .4459/3 & 72.68/6 & 52.51/6 & .6097/6 & .4451/6 \\
    LLaVA-1.5-13B & 64.97/10 & 37.24/11 & .4734/10 & .2397/10 & 68.49/10 & 47.32/8 & .5597/8 & .3668/9 \\
    LLaVA-1.6-13B & 63.71/11 & 37.37/10 & .4711/11 & .2412/9 & 69.16/9 & 46.98/9 & .5596/9 & .3646/10 \\
    Qwen-VL-7B-Chat & 67.37/9 & 38.70/9 & .4916/9 & .1630/11 & 66.70/11 & 41.51/11 & .5117/11 & .2188/11 \\
    Qwen2.0-VL-7B & 73.32/7 & 52.92/7 & .6147/7 & .3873/8 & 82.29/1 & 66.93/2 & .7382/2 & .5042/2 \\
    Qwen2.5-VL-7B & 78.70/2 & 61.08/2 & .6878/2 & .4343/4 & 81.23/3 & 65.42/4 & .7247/4 & .4698/5 \\
    Qwen3.0-VL-8B & 75.99/4 & 57.07/4 & .6518/4 & .4324/5 & 82.09/2 & 67.88/1 & .7431/1 & .5026/3 \\
    \midrule    
    Pearson's $r$ & .8863 & .9280 & .9240 & & .8373 & .8388 & .8449 & \\
    Spearman's $\rho$ & .7909 & .8000 & .7909 & & .9182 & .9273 & .9273 & \\
    Kendall's $\tau$ & .6364 & .6727 & .6364 & & .8182 & .7818 & .7818 & \\
    \bottomrule
  \end{tabular}
\end{table}
\noindent {\bf Results on NaturalBench and NatConBench}: The results on NaturalBench and NatConBench are presented in Table~\ref{tab:NaturalBench-NatConBench}. Different from the MC-VQA formats, in NaturalBench, the default accuracy (Acc) is obtained on the individual YN tests on alternating image-question pairs. The baseline accuracy rate on random choice is 50\%. Hence, the Acc score of 70\% to 80\% does not indicate that it is an easy task. When the joint tests on both logical sufficiency and necessary as described in Sec~\ref{sec:lcm-nb} is introduced, we observe the large gaps between the LCM scores to those of Acc. Again, this observation is justified by the scores of JYN-Acc (J-Acc in table) when the gt annotation of correct pair is applied in Eq.~(\ref{eq:jsc_acc_naturalbench}). The observations from Table~\ref{tab:NaturalBench-NatConBench} are very close to those on previous tables on MC-VQA benchmarks, which indicates the effectiveness of LCM on NaturalBench format with negative image involved on logic necessary conditions. The comparison of LCM with combined accuracy metrics proposed in NaturalBench are presented in Appendix.

\subsection{Ablation studies}

\begin{figure}[t]
  \centering
\includegraphics[width=0.8\textwidth]{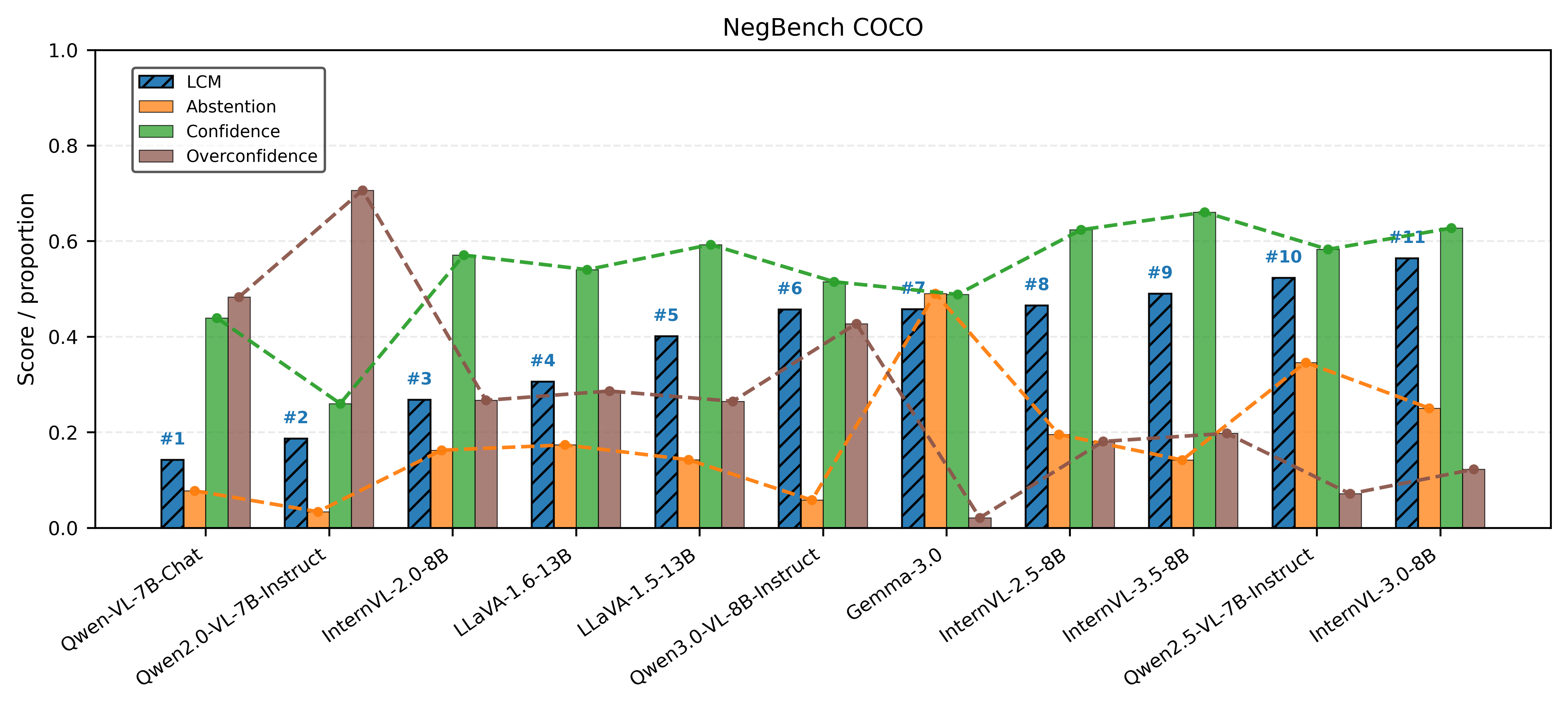} 
\caption{The relation of LCM with response distribution, where the blue bar represents LCM score, and the bars of orange, green and brown colors represent the percentages of Abstention, Confidence and Overconfidence responses.}
\label{fig:lcm-res-dist}
\end{figure}
\noindent {\bf Relation between LCM with YN answer distribution}: 
To better understand why joint YN tests on logical sufficiency and necessity result in low J-Acc and LCM scores despite high MC-VQA accuracy, we conduct a statistical analysis of MLLM response patterns on these joint YN tests.
On recent studies~\cite{Kalai:nature2026}, the response types can be categorized as three types. They are Abstention ({\em i.e.}, no choice is confirmed), Confidence ({\em i.e.}, only one choice is selected), and Overconfidence ({\em i.e.}, more than one choice are selected). Overconfidence may lead to high risk with many hallucinations, and Abstention may be low risk to lead to user's false decisions. We produce a bar graph to visualize the relation of LCM with the response distribution. One representative graph on NegBench COCO is shown in Figure~\ref{fig:lcm-res-dist}, and all the six bar graphs are presented in Appendix. In the figure, the models are sorted on increasing LCM scores (blue bars) from left to right. One may observe a general trend on the bar graph. The increase of LCM score may be related to the reduction of Overconfidence rate and the increasing of Confidence rate, while Abstention rate fluctuates. This observation indicates that a high LCM score might imply a low risk of hallucination and high confidence of reliability.

\begin{figure}[t]
  \centering
\includegraphics[width=0.46\textwidth]{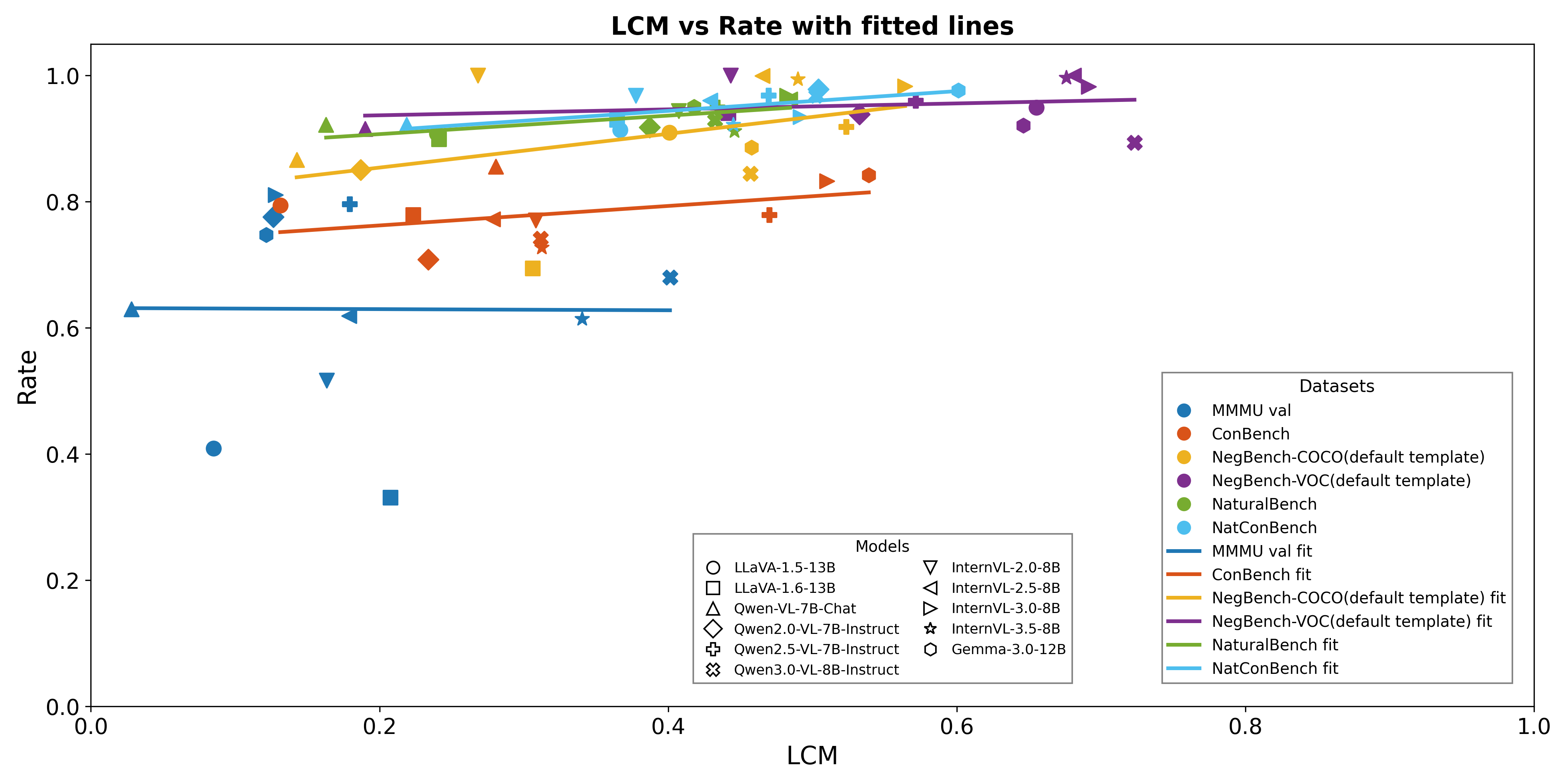} 
\includegraphics[width=0.46\textwidth]{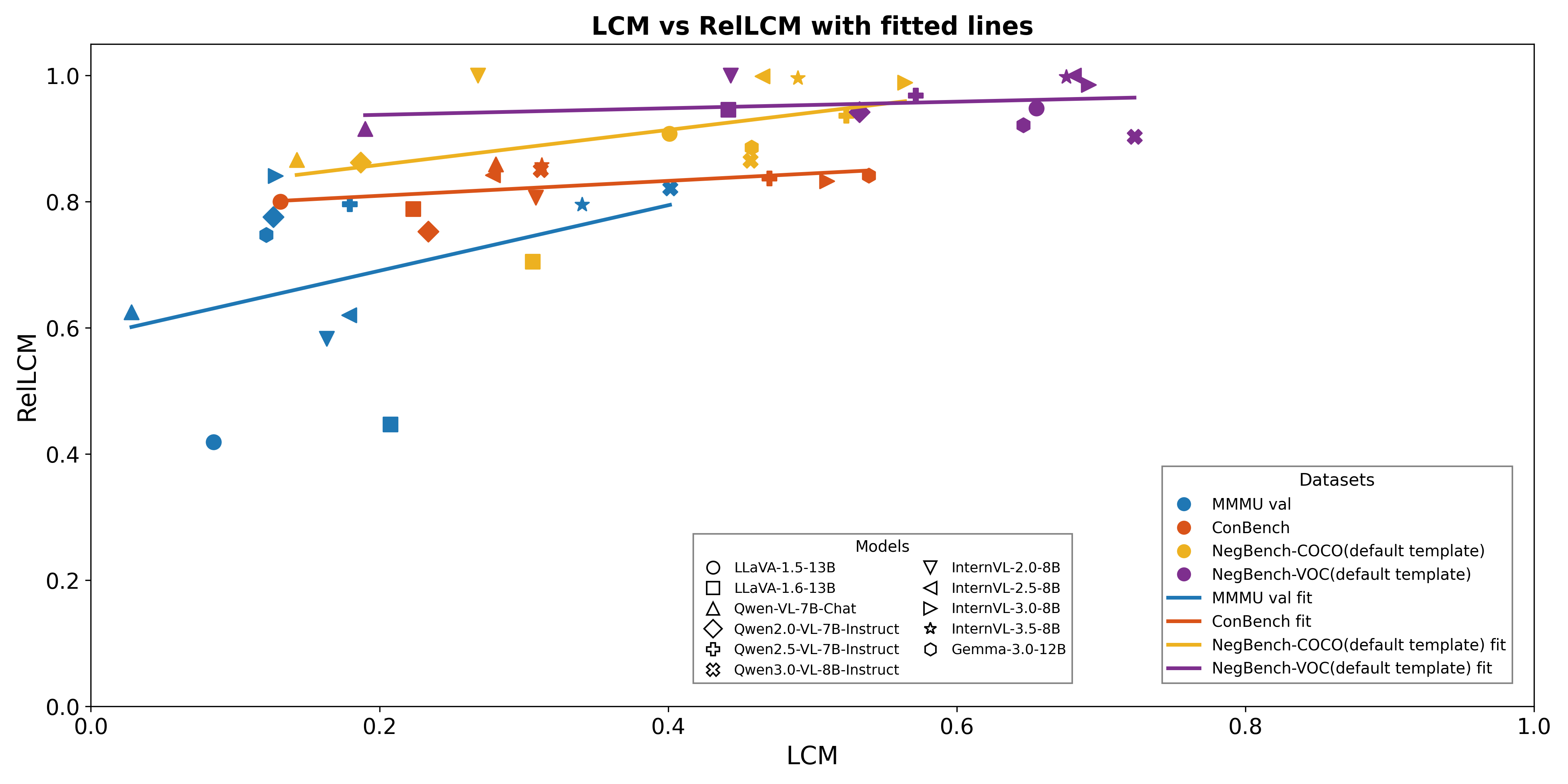} 
\caption{Left: The distribution of ratio $LCM_{gt}/LCM$ wrt LCM scores; Right:  RelLCM vs LCM.}
\label{fig:lcm-rellcm}
\end{figure}
\noindent {\bf Reliability and additional benefits of LCM}:
To evaluate the reliability of LCM, we also compute $LCM_{gt}$ on samples that are both consistent and correct on given ground-truth, as described in Sec~\ref{sec:lcm-mcvqa} and Sec~\ref{sec:lcm-nb}. The ratio $LCM_{gt}/LCM$ is computed for every model across all benchmarks. The distribution of the ratio values wrt LCM scores is shown in the left in Figure~\ref{fig:lcm-rellcm}. It is observed that when LCM>0.3, most ratio values are over 80\%, the mean value is 90.6\%, and the higher the LCM score, the larger the ratio, indicating that higher LCM leads to higher confidence of logic consistency. The LCM is a per-sample measurement. It may be used to justify correct answer even without gt annotation. To validate this assumption, we select consistent answers on large LCM with both $P_{MC}$ and $P_{JYN}$ being larger than 0.5. Let the number of such reliable answers be $N_R$. We also count the number of such samples which match the gt labels, $N_{Rgt}$. The ratio of $N_{Rgt}/N_R$ indicates the percentage of true correct answers wrt selected reliable answers on LCM. The distribution of the ratio wrt LCM scores for every model across all benchmarks is shown in the right in Figure~\ref{fig:lcm-rellcm}. One can find almost the same observations on the left plot, which indicates that we can use LCM to select correct answer with over 80\% of accuracy when LCM>0.3.

\subsection{Summary of new findings and limitations}

\noindent {\bf Finding 1}: While the frontier open-source MLLMs achieve rapid progress in accuracy levels on public benchmarks, their performances on logic consistency are still low, and lag far behind accuracy. Beyond accuracy, logic consistency might be able to provide a deeper insight on both accuracy and reliability, without the need of additional dataset and/or annotation. {\bf Finding 2}: VL-LCM is strongly correlated with F1 of MC-Acc and JYN-Acc, revealing its effectiveness for annotation-free model ranking, selection and validation on novel tasks. {\bf Finding 3}: As per-sample measurement, it could be used for correct answer selection and justification without gt annotation in online application, such as natural human-agent interaction.

\noindent {\bf Limitations}: There are still many open problems for future studies, such as that human language and natural vision-language relations might not be exactly yes or no logically, the consistent answer from MC and JYN tests may not be the right answer due to biased model, additional YN tests naturally introduce increased computational cost (see App), how to extend to open-end natural language answers, how to exploit VL-LCM to improve MLLM's reliability  and deploy it in applications.

\section{Conclusions}
\label{sec:conclusion}

To validate the logic consistency of MLLM despite high accuracy, even without gt annotation in novel applications, we propose a novel framework to evaluate MLLM on both logical sufficient and necessary conditions, and derive a new metric, {\em i.e.}, Vision-Language Logic Consistency Metric (VL-LCM) for typical MC-VQA and recent NaturalBench formats. We perform comprehensive experiments and extensive evaluations on five recent challenging VL Benchmarks with 11 recent open-source MLLMs from 4 frontier families. Our new findings reveal the effectiveness of VL-LCM for evaluation on both accuracy and reliability beyond accuracy, for annotation-free evaluation of MLLM on model ranking, selection and validation for new tasks, and reliable answer selection and justification without gt annotation. The limitations and open problems are discussed for suggestions of future studies and potential applications.





{
\small

    \bibliographystyle{ieeenat_fullname}
    \bibliography{main}

}


\appendix

\section{VL Benchmarks}
\noindent {\bf NegBench} is a recent benchmark designed to assess visual–language understanding under factual and counter‑factual reasoning. Each sample contains both positive and negative phrases, and choices follow three linguistic templates: Affirmation, Negation, and Hybrid. An affirmation text contains only positive elements $\{pos\}$, i.e., objects present in the image. A negation text includes only negative elements $\{ neg \}$, i.e., objects absent from the image but commonly associated with the present objects. A hybrid text contains both positive and negative elements. There are three natural image MCQ tasks, {\em i.e.}, COCO, VOC2007, and HardNeg-Syn. Our experiments are conducted on the publicly available data, i.e., 5914 MCQ questions on COCO and 5032 MCQ questions on VOC2007.

\noindent {\bf ConBench} is a recent multimodal benchmark designed to systematically assess the consistency of large Vision-Language Models (LVLMs). It focuses on identifying inconsistencies in model responses when different question formats are applied from the same underlying knowledge point. The full benchmark comprises 1,000 public images and a total of 4,000 questions (including 3,000 discriminative ground truths) spanning 19 distinct categories. Each image case includes four prompts centered on a single knowledge point: three discriminative prompts (i.e., True/False, multiple-choice, limited VQA) and one generative caption prompt. ConBench hierarchically evaluates LVLMs across three core capabilities: Sensation, Cognition, and Knowledge, corresponding to Easy, Medium and Hard Modes of difficulty. In our experiments, we utilize the questions of multiple-choice formats. We obtain a subset of 1,009 MC-VQA samples for logic consistency analysis.

\noindent {\bf MMMU} is a representative benchmark to evaluate multi-modal models for real-world reasoning tasks, and employed in leaderboards for ranking the latest frontier MLLMs~\cite{mmmu_leaderboard}. Comprising comprehensive 11.5K multimodal questions across six core disciplines ({\em i.e.}, Art \& Design, Business, Science, Health \& Medicine, Humanities \& Social Science, and Tech \& Engineering) over 30 highly heterogeneous image types (e.g. charts, diagrams, maps, tables, music sheets, and chemical structures), MMMU focuses on advanced visual perception, reasoning ({\em e.g.}, logical, spatial, commonsense, mathematical) and knowledge ({\em e.g.} domain expertise linguistic, world). MMMU is designed to test models beyond simple perception, requiring integration of visual and textual information, application of domain-specific understanding, and demonstration of advanced reasoning skills. In our experiments to evaluate the effectiveness of VL-LCM, we perform the experiments on the val set, as the gt annotation of test set is not available publicly. 

\noindent {\bf NaturalBench} is a new benchmark designed for reliably evaluating VLMs using natural adversarial samples, focusing on questions about natural images that are straightforward for humans but challenging for state-of-the-art models. A crucial feature of NaturalBench is its vision-centric design where each question is paired with two images that yield different answers, enforcing VLMs to rely on visual input and preventing ``blind'' solutions that exploit language priors.  The samples are collected via a semi-automated pipeline from natural image-text datasets, which include yes/no and multiple-choice formats. The benchmark is challenging due to its focus on compositionality that requires diverse visio-linguistic skills (e.g., Object, Attribute, Relation, Reasoning). 

\section{Creation of NatConBench}

We introduce NatConBench, a new dataset constructed from existing ConBench, where images and texts are paired across 19 diverse categories. Under each category, we randomly select two questions on two different images, where each question has one correct answer on one image alternatively. They form four image-sentence pairs, with two ``yes'' and two ``no'' answers:
\\ \noindent $\bullet$ $V_1$, $T_1$ : yes
\\ \noindent $\bullet$ $V_1$, $T_2$ : no
\\ \noindent $\bullet$ $V_2$, $T_1$ : no
\\ \noindent $\bullet$ $V_2$, $T_2$ : yes

 The process is detailed as follows: \textbf{1) Multiple-Choice Pairs}: We randomly sampled two existing multiple-choice samples from the ConBench dataset. A pair is retained if the images and their corresponding ground-truth answers are different. The paired samples are allowed to share the same question text, where the difference in ground truth answers ensures the vision-centric adversarial of the NaturalBench protocol. \textbf{2) True/False Pairs}: We randomly selected two existing True/False samples from ConBench dataset. A pair is retained if the images and questions are different, but the ground-truth answers are the same (i.e., both are `Yes'). This configuration evaluates model consistency with varied visual and linguistic context under the same knowledge point, further capturing model biases as exposed in NaturalBench.
 
\begin{figure}[h]
  \centering
\includegraphics[width=1.0\textwidth]{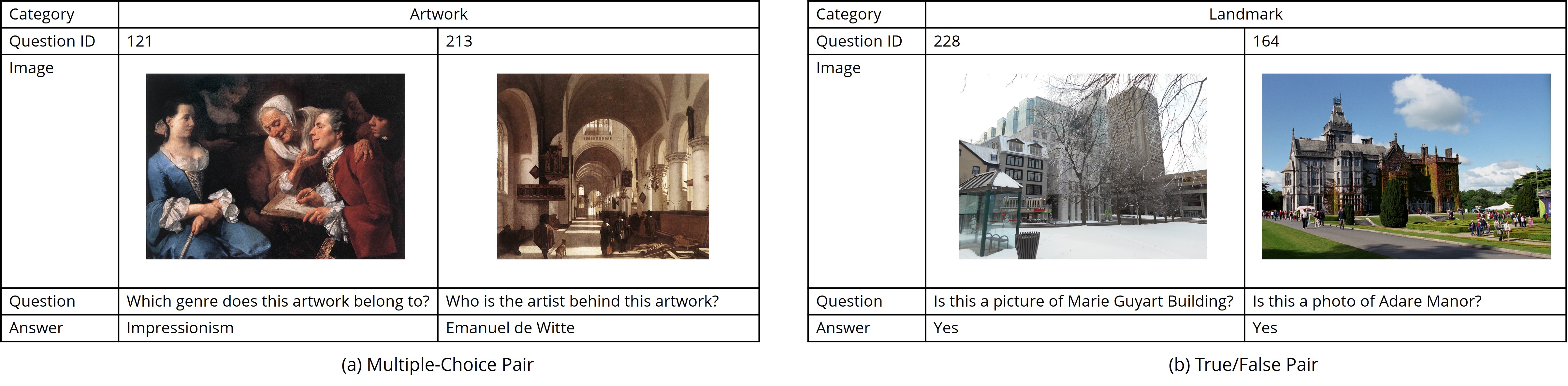} 
\caption{Selected source image-question pairs from ConBench for creation of NatConBench.}
\label{fig:conbench_samples}
\end{figure}

\begin{figure}[h]
  \centering
\includegraphics[width=1.00\textwidth]{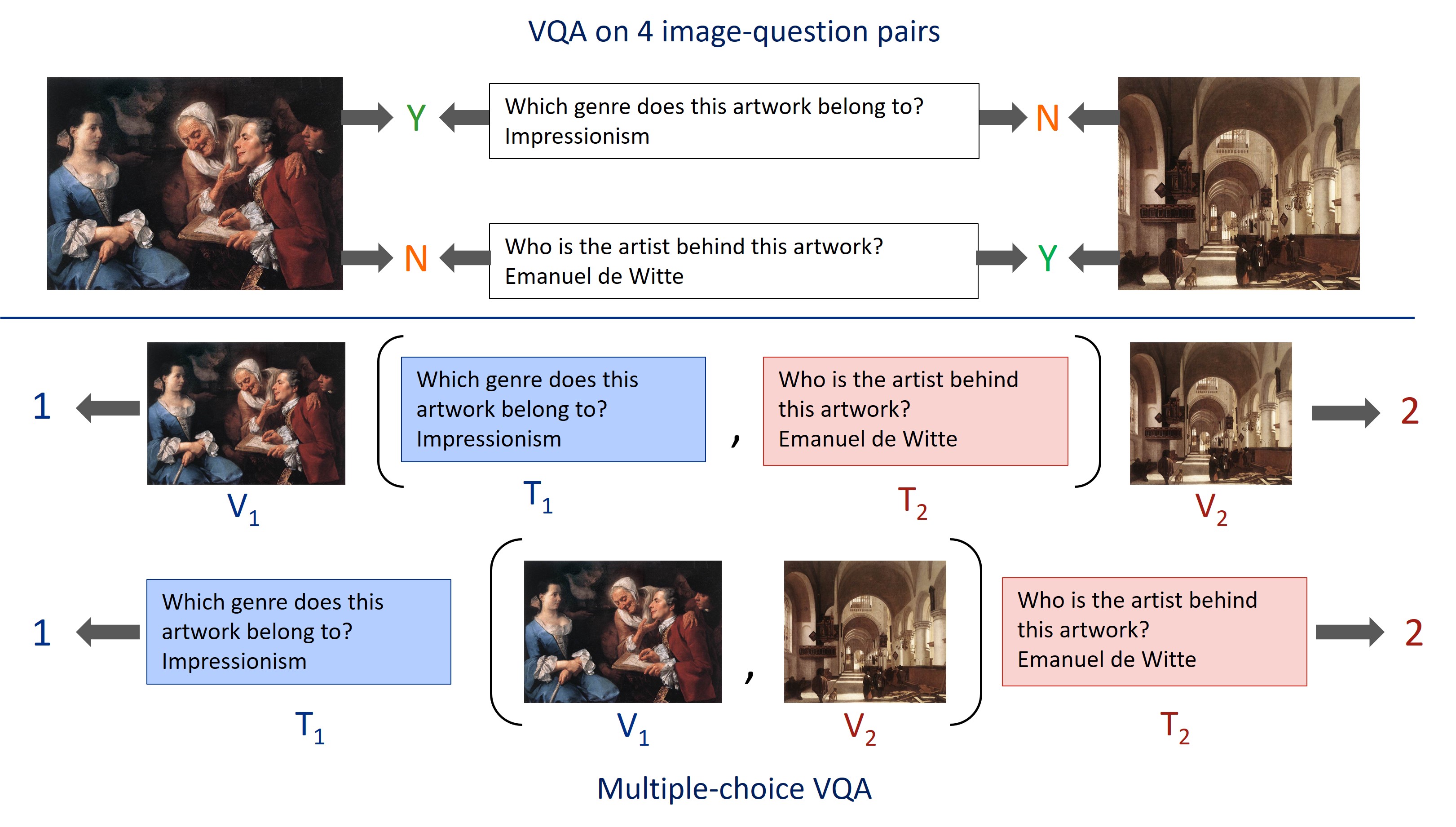} 
\caption{The example of the generated NatConBench sample on the source Multiple-Choice pair samples shown in~\ref{fig:conbench_samples}(a).}
\label{fig:natconbench_mc}
  \centering
\includegraphics[width=1.00\textwidth]{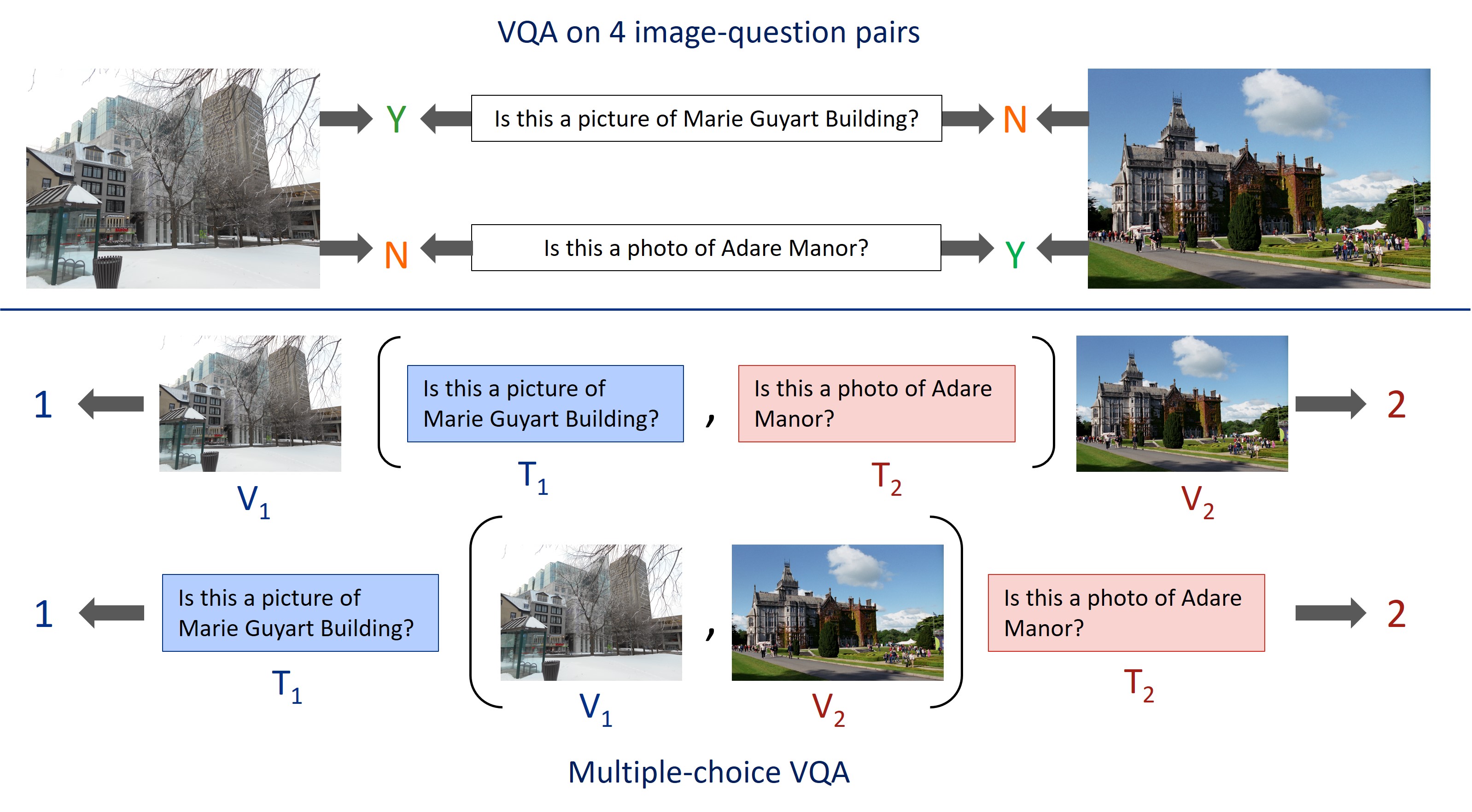} 
\caption{The example of the generated NatConBench sample on the source True/False pair samples shown in~\ref{fig:conbench_samples}(b).}
\label{fig:natconbench_tf}
\end{figure}

For illustration, two source samples of image-question pairs selected from ConBench are presented in~\ref{fig:conbench_samples}. The NaturalBench style sample generated on~\ref{fig:conbench_samples}(a) is presented in~\ref{fig:natconbench_mc}, and the NaturalBench style sample generated on ~\ref{fig:conbench_samples}(b) is displayed in~\ref{fig:natconbench_tf}, respectively.

\section{Prompting strategies}
\begin{figure}[h]
  \centering
\includegraphics[width=0.7\columnwidth]{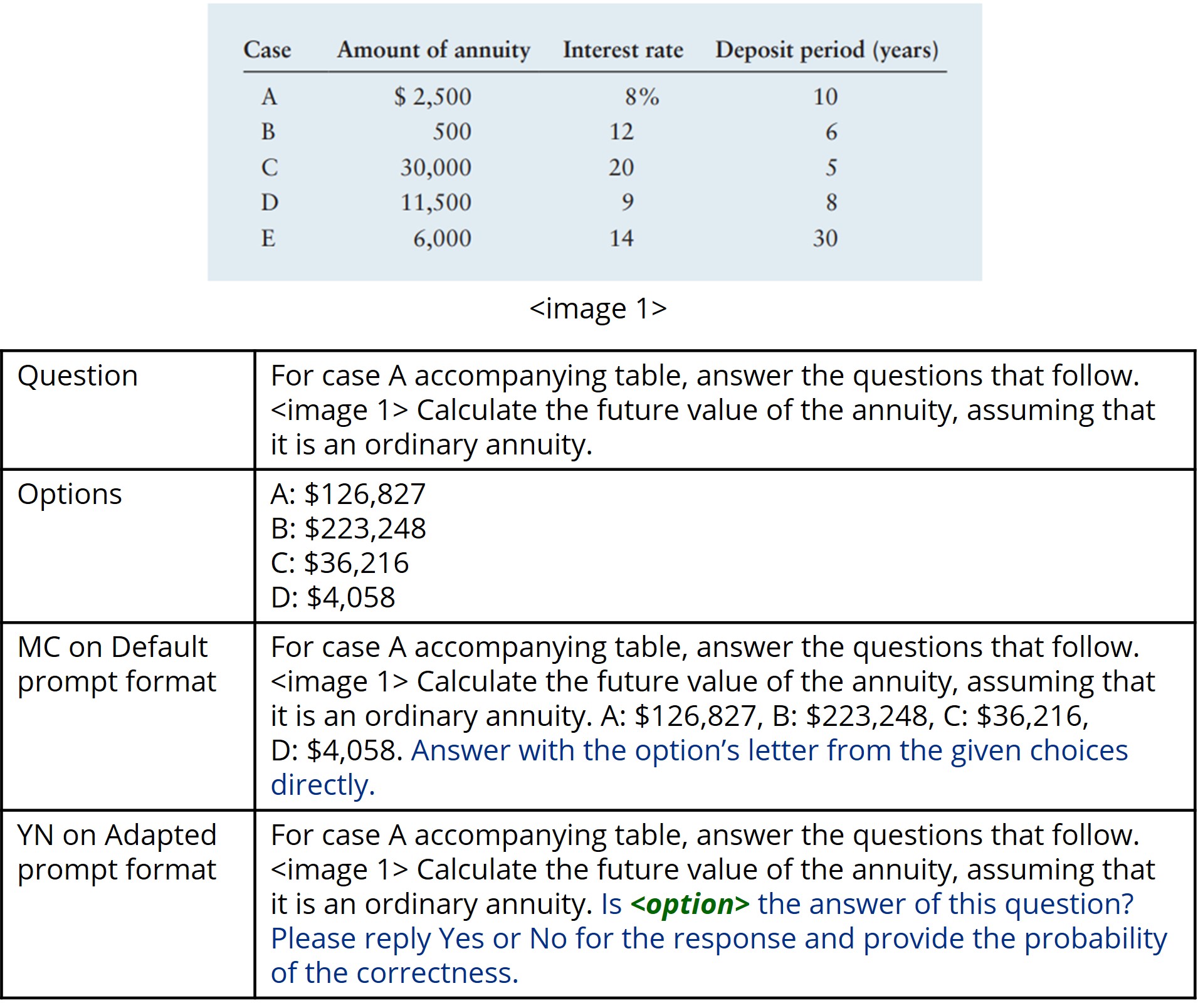} 
\caption{Examples of prompts on one sample from MMMU.}
\label{fig:prompt_types_mmmu}
\end{figure}

\begin{figure}[h]
  \centering
\includegraphics[width=1.0\columnwidth]{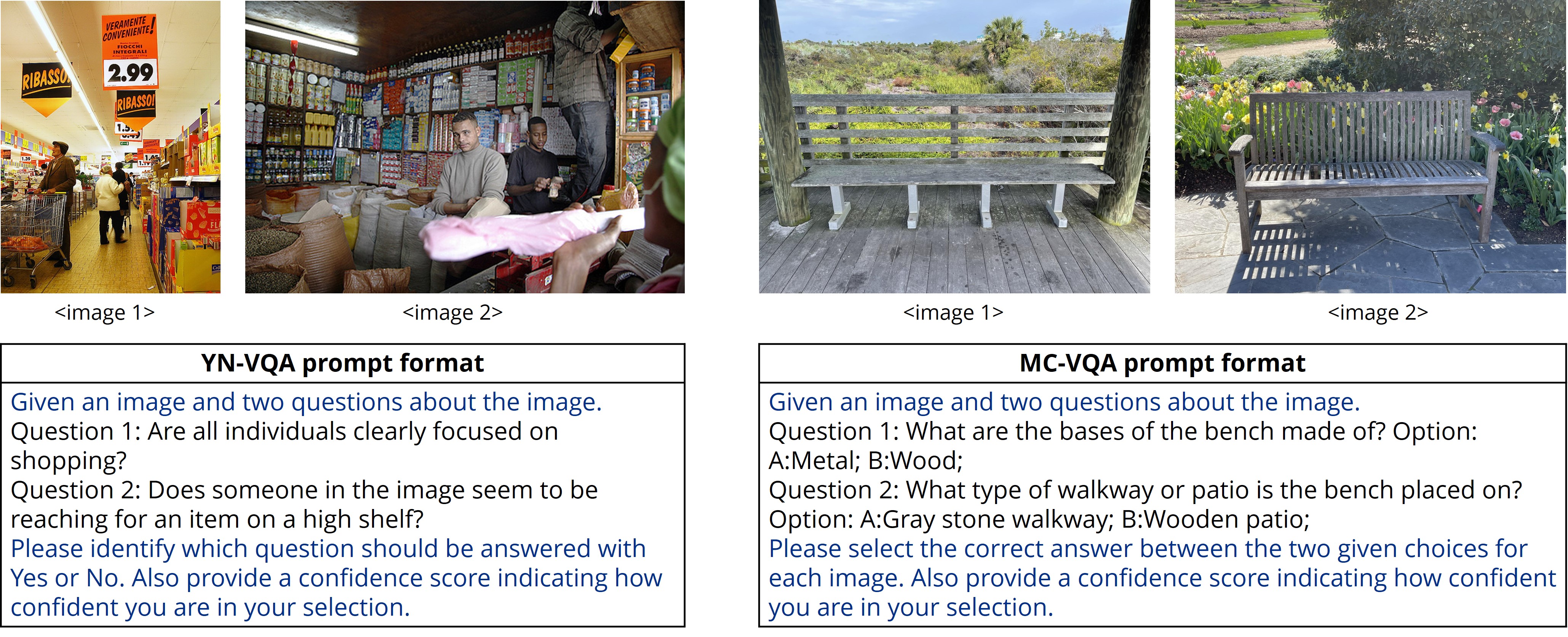} 
\caption{Examples of prompts on one sample from NaturalBench.}
\label{fig:prompt_types_naturalbench}
\end{figure}

In each public benchmark for vision-language understanding on VQA tasks, such as MMMU which is widely employed for frontier MLLM leader-board, there are provided prompt templates of VQA for training and evaluation. 
As the purpose of this study is to investigate the logic consistency to different multimodal inputs under both sufficient and necessary conditions, in the experiments reported in the main paper, we employ the provided default prompt for the main test, and a proposed variant prompt for another test. 
Specifically, on MMMU and ConBench, we use the default prompt template for the Multi-Choice (MC) test ({\em i.e.}, MC with default prompt), 
and the variant prompt template for the Yes-No (YN) test ({\em i.e.}, YN with adapted prompt), which is adapted from the default prompt template.
Examples of both MC and YN prompt formats for MMMU and NaturalBench are presented in Figures~\ref{fig:prompt_types_mmmu} and ~\ref{fig:prompt_types_naturalbench} respectively. Recent studies of prompt skills are considered for stable performance~\cite{ismithdeen-etal-2025-promptception,mohanty2025futuremllmpromptingadaptive}.

\section{Full experimental results}

The full results on ConBench are presented in Table~\ref{tab:ConBenchFull}, which includes the detailed results on categories of Sensation, Cognition, and Knowledge. It can be found that the performance on the three categories are close to the overall performance, indicating the consistency of LCM metric on different categories. 
\begin{table}
  \caption{Full results on ConBench (Categories).}
  \label{tab:ConBenchFull}
  \footnotesize
  \centering
  \begin{tabular}{l|cc|cc|cc|cc}
    \toprule
     & \multicolumn{2}{c|}{Overall} & \multicolumn{2}{c|}{Sensation} & \multicolumn{2}{c|}{Cognition} & \multicolumn{2}{c}{Knowledge} \\
    \cmidrule(r){2-9}
    MLLM & Acc/J-Acc/F1 & LCM & Acc/J-Acc/F1 & LCM & Acc/J-Acc/F1 & LCM & Acc/J-Acc/F1 & LCM \\
    \midrule
    Gemma3.0-12B & 70.8/46.2/.559 & .539 & 74.3/50.2/.599 & .614 & 66.9/36.5/.472 & .441 & 67.6/48.4/.564 & .492 \\
    InternVL-2.0-8B & 64.5/28.1/.391 & .308 & 67.7/29.9/.415 & .318 & 63.1/37.6/.472 & .407 & 59.4/13.9/.226 & .181 \\
    InternVL-2.5-8B & 72.1/21.8/.335 & .279 & 74.7/22.7/.348 & .279 & 71.5/22.4/.348 & .297 & 67.2/19.3/.299 & .258 \\
    InternVL-3.0-8B & 71.5/44.2/.546 & .510 & 73.1/49.4/.590 & .576 & 71.1/42.6/.533 & .501 & 68.4/35.3/.465 & .385 \\
    InternVL-3.5-8B & 74.5/18.8/.301 & .312 & 76.9/18.9/.304 & .313 & 72.2/25.5/.377 & .361 & 72.1/11.5/.198 & .259 \\
    LLaVA-1.5-13B & 53.0/13.4/.214 & .131 & 56.6/19.7/.293 & .185 & 51.0/7.2/.127 & .077 & 48.0/7.0/.122 & .079 \\
    LLaVA-1.6-13B & 46.8/27.3/.344 & .223 & 49.4/28.9/.365 & .260 & 47.2/22.1/.301 & .176 & 41.0/29.5/.343 & .198 \\
    Qwen-VL-7B-Chat & 56.9/28.8/.383 & .281 & 57.8/37.7/.456 & .366 & 50.6/23.2/.318 & .219 & 61.9/16.8/.264 & .172 \\
    Qwen2.0-VL-7B & 71.8/16.6/.269 & .234 & 74.7/15.9/.263 & .231 & 66.5/20.2/.309 & .272 & 71.3/13.9/.233 & .199 \\
    Qwen2.5-VL-7B & 73.5/39.2/.511 & .470 & 75.5/43.6/.553 & .506 & 70.0/39.2/.502 & .471 & 73.4/29.9/.425 & .397 \\
    Qwen3.0-VL-8B & 76.5/21.4/.335 & .312 & 77.9/24.1/.368 & .339 & 73.0/21.3/.330 & .313 & 77.5/16.0/.265 & .254 \\
    \bottomrule
  \end{tabular}
\end{table}

The full results on MMMU val set are presented in Table~\ref{tab:MMMU-val-full}, which includes the detailed results on six core disciplines, {\em i.e.}, Art \& Design, Business, Science, Health \& Medicine, Humanities \& Social Science, and Tech \& Engineering. In the table, the MC-ACC/JYN-Acc are presented in each column. From the results, one can find different difficulties of the discipline subsets. The disciplines of Business, Science and Tech \& Eng seem to be much challenging than Art \& Design and Human \& Social Sci. However, for all the disciplines, we observe the large drops of JYN-Acc scores compared to the corresponding MC-Acc scores, matching the observations on the overall performance described in main paper without exception. This observation indicates the consistency and effectiveness of logic consistency evaluation on subsets of different difficulties. 
\begin{table}
  \caption{Full results on MMMU val set.}
  \label{tab:MMMU-val-full}
  \footnotesize
  \centering
  \begin{tabular}{l|ccccccc|c}
    \toprule
    MLLM & Overall & Art \& & Business & Science & Health \& & Human \& & Tech \& & LCM \\
     &  & Design &  &  & Medicine & Social Sci. & Eng. & \\
    \midrule
    Gemma3.0-12B & 51.4/10.6 & 60.0/16.7 & 49.3/10.5 & 43.1/7.3 & 47.6/11.7 & 71.4/21.0 & 43.8/2.1 & .122\\
    InternVL-2.0-8B & 45.5/15.7 & 57.5/19.2 & 43.3/14.9 & 32.9/15.3 & 46.9/16.6 & 59.7/18.5 & 38.5/12.0 & .163 \\
    InternVL-2.5-8B & 50.7/18.0 & 63.3/21.7 & 46.3/18.7 & 33.6/11.7 & 57.2/20.0 & 66.4/26.9 & 43.2/19.4 & .179 \\
    InternVL-3.0-8B & 55.4/12.5 & 69.2/17.5 & 50.0/8.2 & 41.6/10.2 & 62.8/14.5 & 73.1/26.9 & 43.8/3.7 & .128 \\
    InternVL-3.5-8B & 57.7/26.5 & 70.8/27.5 & 53.7/26.1 & 51.1/27.7 & 64.8/36.6 & 74.8/30.3 & 41.2/15.1 & .340 \\
    LLaVA-1.5-13B & 38.3/8.4 & 50.8/5.0 & 25.4/12.7 & 29.9/7.3 & 40.0/6.9 & 54.6/5.9 & 33.9/10.9 & .085 \\
    LLaVA-1.6-13B & 37.1/16.1 & 54.2/10.0 & 27.6/22.4 & 26.3/13.1 & 35.2/13.1 & 53.8/16.8 & 31.8/19.3 & .208 \\
    Qwen-VL-7B-Chat & 37.9/3.5 & 50.8/6.7 & 35.1/3.7 & 33.6/2.2 & 40.0/0.7 & 42.0/3.4 & 30.7/4.7 & .028 \\
    Qwen2.0-VL-7B & 52.5/11.5 & 68.3/17.5 & 48.5/9.7 & 39.4/11.0 & 53.8/11.7 & 68.9/19.3 & 43.8/4.2 & .126\\
    Qwen2.5-VL-7B & 54.1/14.5 & 68.3/31.7 & 46.3/6.0 & 43.1/13.1 & 55.9/11.7 & 67.2/29.4 & 49.0/3.6 & .180 \\
    Qwen3.0-VL-8B & 54.9/22.9 & 69.2/40.0 & 50.8/14.2 & 44.5/17.5 & 60.0/31.0 & 65.6/39.5 & 45.8/5.7 & .401 \\
    \bottomrule
  \end{tabular}
\end{table}

The full results on NaturalBench are presented in Table~\ref{tab:NaturalBench-full}, where the results of combined accuracies proposed in the Benchmark are presented. Three metrics are introduced in NaturalBench: \textbf{1) Question Accuracy (Q-Acc)}: A point is given if the model correctly answers a question for both paired images, \textbf{2) Image Accuracy (I-Acc)}: A point is given when the model correctly answers both questions associated with a single image, \textbf{3) Group Accuracy (G-Acc)}: A point is given only if the model correctly answers all four QA pairs within a test case. One can find the clear drops of scores from Acc, to Q-Acc and I-Acc, to G-Acc, but these drops might not give a right understanding, as the baselines on random choices are different. The rate of random choice for Acc is 50\%, the one for Q-Acc and I-Acc is 25\%, and that for G-Acc is 12.5\%. In addition, they are all obtained on gt annotation under logic sufficient conditions. VL-LCM is more strongly correlated to the F1 of Acc and J-Acc, represents the accuracy and reliability on both logic sufficient and necessary conditions.
\begin{table}
  \caption{Full results on NaturalBench.}
  \label{tab:NaturalBench-full}
  \small
  \centering
  \begin{tabular}{l|cccccc|c}
    \toprule
    MLLM & Acc & J-Acc & F1 & Q-Acc & I-Acc & G-Acc & LCM \\
    \midrule
    Gemma3.0-12B & 74.84 & 54.53 & .6309 & .5195 & .5487 & .2584 & .4179 \\
    InternVL-2.0-8B & 74.22 & 53.87 & .6243 & .5029 & .5471 & .2337 & .4074 \\
    InternVL-2.5-8B & 78.70 & 61.08 & .6878 & .5882 & .6139 & .3274 & .4845 \\
    InternVL-3.0-8B & 79.80 & 62.92 & .7036 & .6079 & .6318 & .3621 & .4825 \\
    InternVL-3.5-8B & 71.41 & 49.71 & .5862 & .4745 & .4803 & .1816 & .4459 \\
    LLaVA-1.5-13B & 64.97 & 37.24 & .4734 & .3424 & .4008 & .1200 & .2397 \\
    LLaVA-1.6-13B & 63.71 & 37.37 & .4711 & .3432 & .4005 & .1263 & .2412 \\
    Qwen-VL-7B-Chat & 67.37 & 38.70 & .4916 & .3608 & .3934 & .1274 & .1630 \\
    Qwen2.0-VL-7B & 73.32 & 52.92 & .6147 & .5058 & .5095 & .2037 & .3873 \\
    Qwen2.5-VL-7B & 78.70 & 61.08 & .6878 & .5955 & .6142 & .3358 & .4343 \\
    Qwen3.0-VL-8B & 75.99 & 57.07 & .6518 & .5532 & .5524 & .2526 & .4324 \\
    \bottomrule
  \end{tabular}
\end{table}

The full results on NatConBench with combined accuracies are presented in Table~\ref{tab:NatConBench-summary}, and the full results with detailed performance on the three categories are presented in Table~\ref{tab:NatConBenchCategory}. On summary, the results are similar to those on NaturalBench reported in Table~\ref{tab:NaturalBench-full}, and on categories, the results are similar to those on ConBench reported in Table~\ref{tab:ConBenchFull}. These observations indicate the consistency of LCM when tested on different formats with both logic sufficient and necessary conditions.
\begin{table}
  \caption{Full results on NatConBench (Summary).}
  \label{tab:NatConBench-summary}
  \small
  \centering
  \begin{tabular}{l|cccccc|c}
    \toprule
    MLLM & Acc & J-Acc & F1 & Q-Acc & I-Acc & G-Acc & LCM \\
    \midrule
    Gemma3.0-12B & 80.98 & 65.92 & .7268 & .6453 & .6592 & .4559 & .6011 \\
    InternVL-2.0-8B & 70.70 & 44.30 & .5447 & .4247 & .4430 & .2525 & .3776 \\
    InternVL-2.5-8B & 72.23 & 48.16 & .5779 & .4620 & .4816 & .2838 & .4290 \\
    InternVL-3.0-8B & 73.85 & 54.41 & .6266 & .5162 & .5441 & .3151 & .4918 \\
    InternVL-3.5-8B & 72.68 & 52.51 & .6097 & .4765 & .5251 & .2006 & .4451 \\
    LLaVA-1.5-13B & 68.49 & 47.32 & .5597 & .4223 & .4732 & .2514 & .3668 \\
    LLaVA-1.6-13B & 69.16 & 46.98 & .5596 & .4330 & .4698 & .2425 & .3646 \\
    Qwen-VL-7B-Chat & 66.70 & 41.51 & .5117 & .3654 & .4011 & .2235 & .2188 \\
    Qwen2.0-VL-7B & 82.29 & 66.93 & .7382 & .6587 & .6693 & .4838 & .5042 \\
    Qwen2.5-VL-7B & 81.23 & 65.42 & .7247 & .6453 & .6542 & .4760 & .4698 \\
    Qwen3.0-VL-8B & 82.09 & 67.88 & .7431 & .6508 & .6788 & .4749 & .5026 \\
    \bottomrule
  \end{tabular}
\end{table}

\begin{table}
  \caption{Full results on NatConBench (Categories).}
  \label{tab:NatConBenchCategory}
  \footnotesize
  \centering
  \begin{tabular}{l|cc|cc|cc|cc}
    \toprule
     & \multicolumn{2}{c|}{Overall} & \multicolumn{2}{c|}{Sensation} & \multicolumn{2}{c|}{Cognition} & \multicolumn{2}{c}{Knowledge} \\
    \cmidrule(r){2-9}
    MLLM & Acc/J-Acc/F1 & LCM & Acc/J-Acc/F1 & LCM & Acc/J-Acc/F1 & LCM & Acc/J-Acc/F1 & LCM \\
    \midrule
    Gemma3.0-12B & 81.0/65.9/.727 & .601 & 90.6/81.9/.860 & .783 & 75.7/55.2/.638 & .477 & 76.3/60.0/.672 & .536 \\
    InternVL-2.0-8B & 70.7/44.3/.545 & .378 & 84.8/70.8/.771 & .637 & 61.3/24.7/.352 & .208 & 65.4/36.2/.466 & .276 \\
    InternVL-2.5-8B & 72.2/48.2/.578 & .429 & 87.0/75.0/.806 & .694 & 64.5/32.5/.433 & .275 & 64.6/35.8/.461 & .307 \\
    InternVL-3.0-8B & 73.9/54.4/.627 & .492 & 87.5/76.1/.814 & .713 & 67.6/42.6/.522 & .356 & 66.0/43.7/.526 & .397 \\
    InternVL-3.5-8B & 72.7/52.5/.610 & .445 & 86.9/75.3/.807 & .646 & 66.0/38.9/.490 & .300 & 64.6/42.3/.511 & .381 \\
    LLaVA-1.5-13B & 68.5/47.3/.560 & .367 & 84.9/71.2/.775 & .601 & 59.7/34.1/.434 & .234 & 60.3/35.7/.448 & .255 \\
    LLaVA-1.6-13B & 69.2/47.0/.560 & .365 & 83.3/68.1/.750 & .578 & 60.8/33.2/.430 & .237 & 62.8/38.7/.479 & .271 \\
    Qwen-VL-7B-Chat & 66.7/41.5/.512 & .219 & 79.3/62.8/.700 & .348 & 59.3/29.1/.390 & .148 & 61.0/31.8/.418 & .138 \\
    Qwen2.0-VL-7B & 82.3/66.9/.738 & .504 & 91.3/83.3/.871 & .639 & 78.9/59.7/.680 & .444 & 76.4/57.2/.654 & .425 \\
    Qwen2.5-VL-7B & 81.2/65.4/.725 & .470 & 91.4/83.0/.870 & .607 & 77.9/59.5/.675 & .417 & 74.0/53.2/.619 & .381 \\
    Qwen3.0-VL-8B & 82.1/67.9/.743 & .503 & 91.1/83.0/.869 & .628 & 78.8/58.0/.695 & .418 & 76.1/58.0/.658 & .456 \\
    \bottomrule
  \end{tabular}
\end{table}

\section{Ablation study on relation between LCM with YN answer distribution} 
\begin{figure}[t]
  \centering
\includegraphics[width=0.48\textwidth]{fig/bar_NegBench_COCO.png} 
\includegraphics[width=0.48\textwidth]{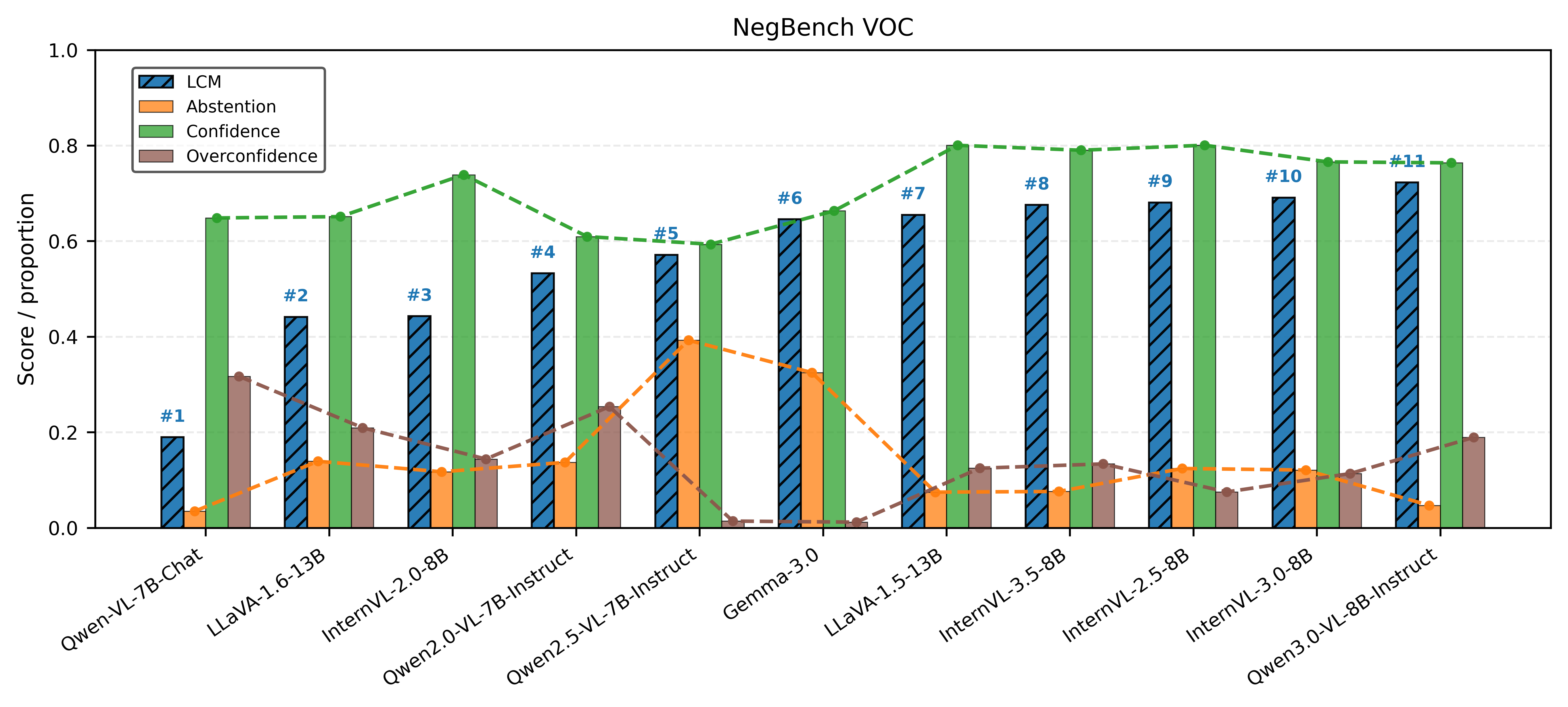} 
\includegraphics[width=0.48\textwidth]{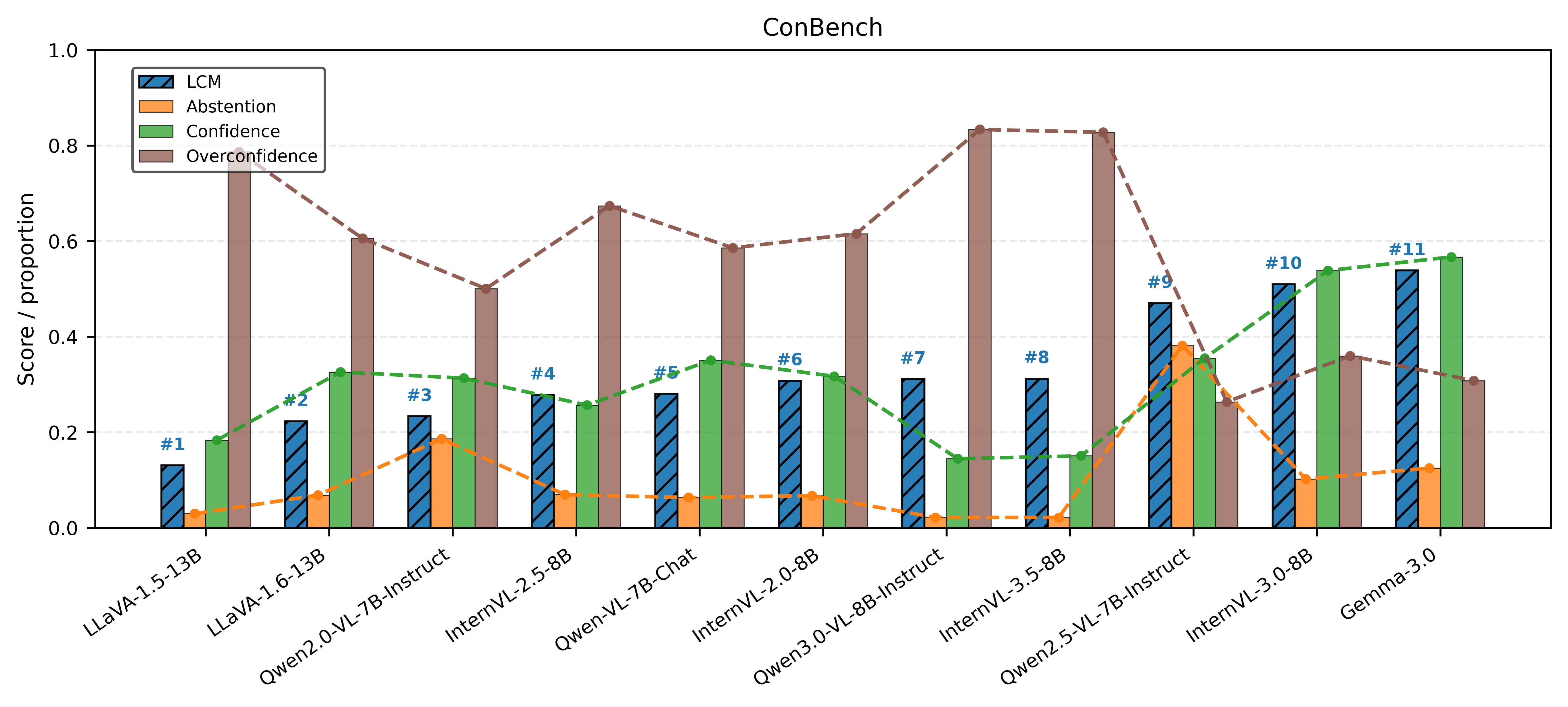} 
\includegraphics[width=0.48\textwidth]{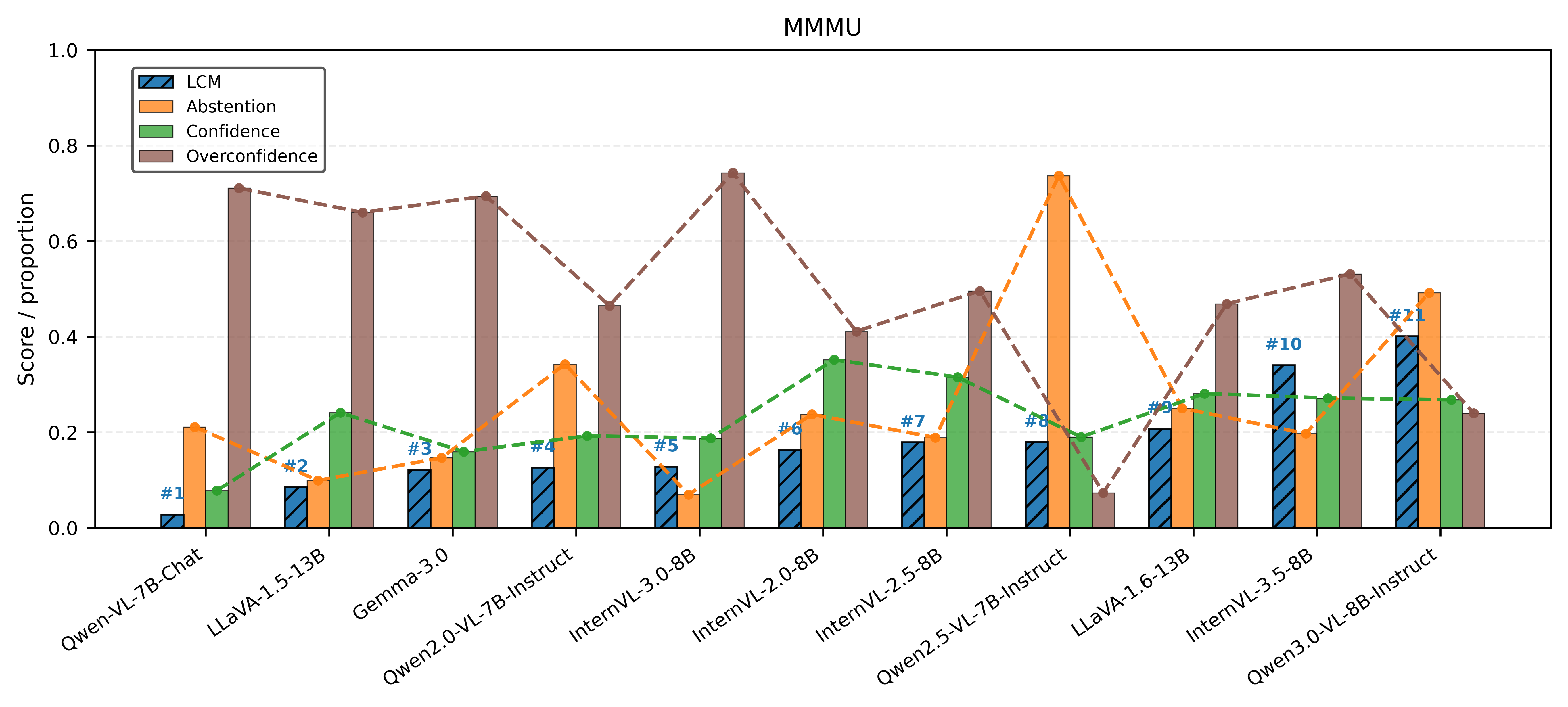} 
\includegraphics[width=0.48\textwidth]{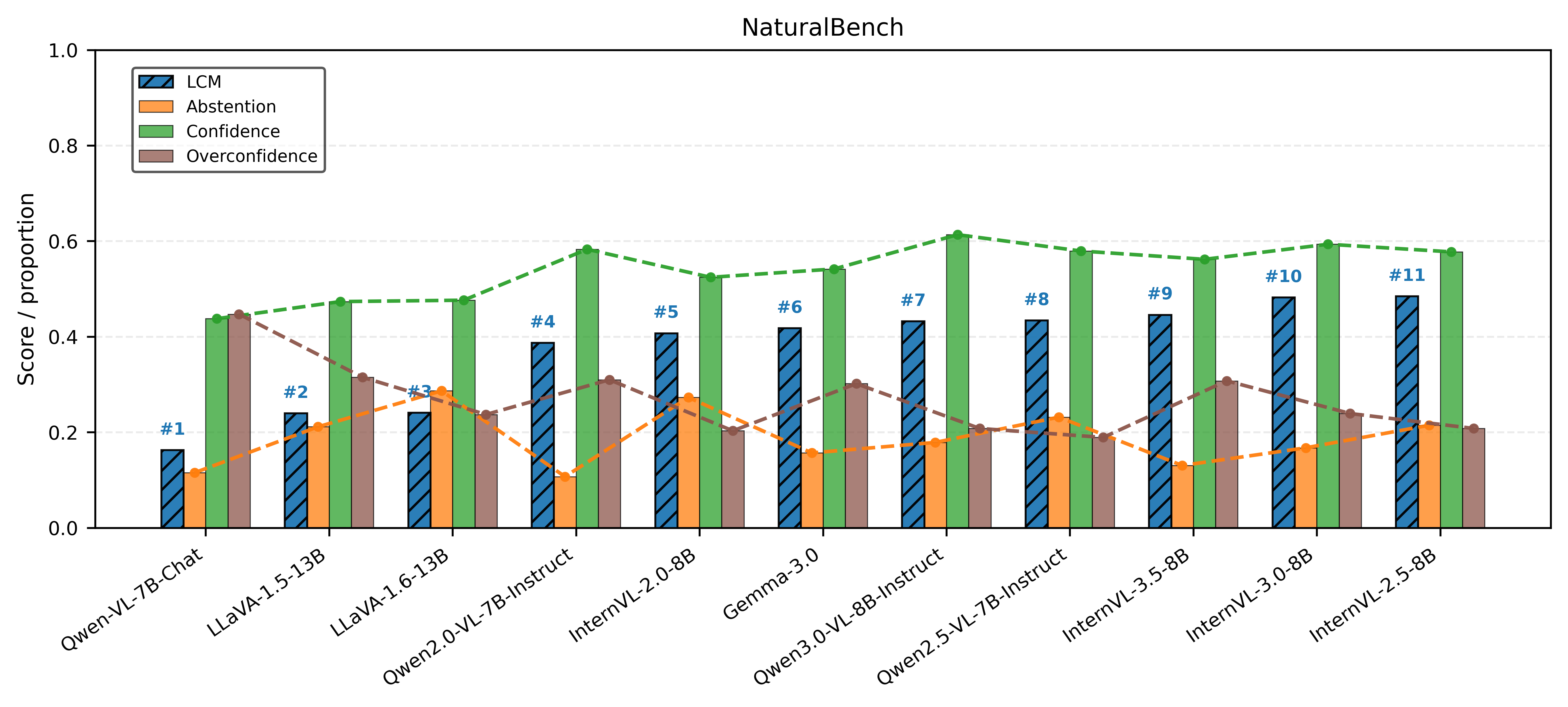} 
\includegraphics[width=0.48\textwidth]{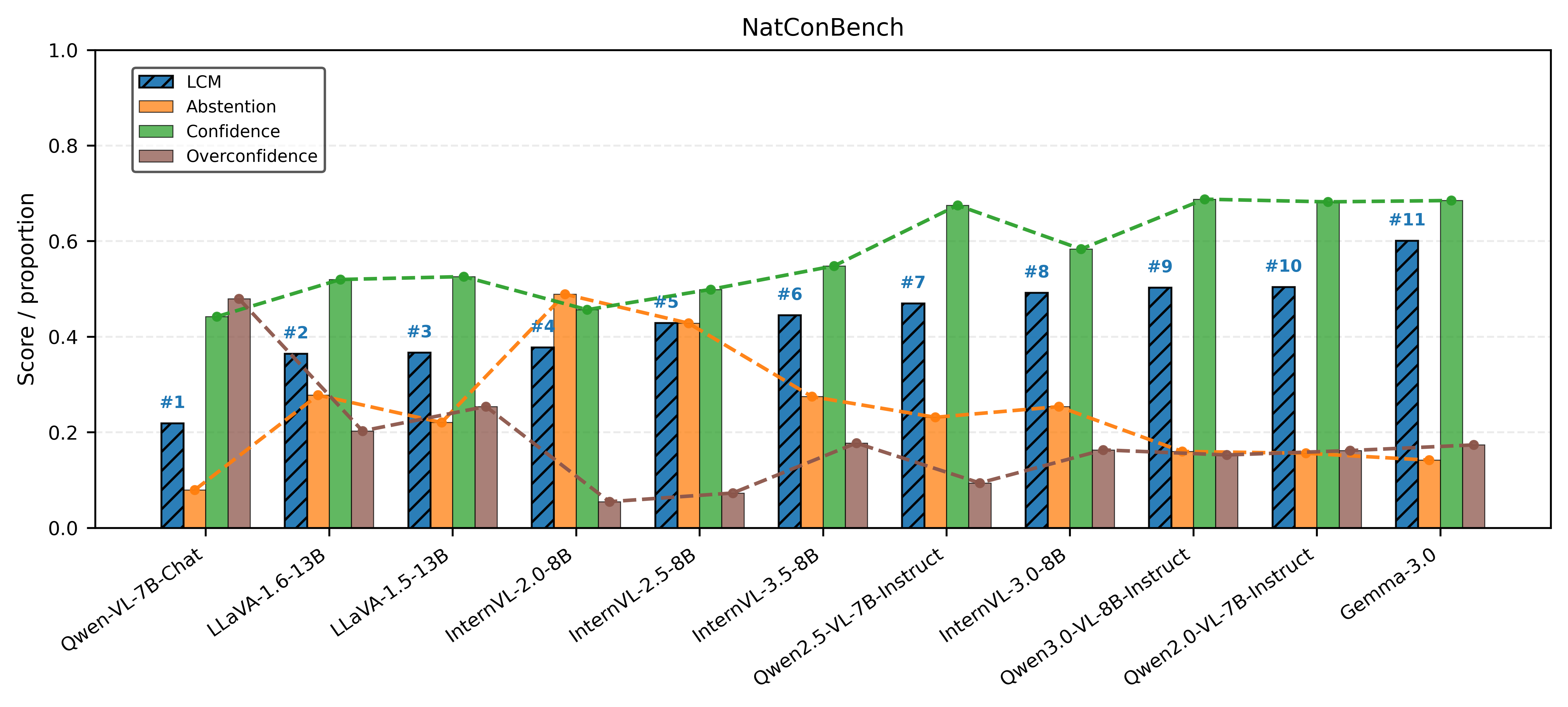} 
\caption{All six bar graphs on the relations of LCM scores with response distributions.}
\label{fig:lcm-res-dist-full}
\end{figure}

All six bar graphs on the statistics of response distributions are presented in Figure~\ref{fig:lcm-res-dist-full}. The general trend, {\em i.e.}, that the increase of LCM score may be related to the reduction of Overconfidence rate and the increasing of Confidence rate while Abstention rate fluctuates, could be observed in all the bar graphs, but it demonstrates with quite large variations with different LCM scores. On NegBench subsets, since the models perform quite well on the traditional CV image sets, the LCM scores are high (with high blue bars). Hence, the rates of Confidence responses (green bars) are high. On the other hand, on the challenging ConBench and MMMU benchmarks, the LCM scores are low, and the rates of Overconfidence (brown bars) are high. In general, from left to right, the green bars increase gradually and brown bars reduce gradually. It can be observed that, to improve the accuracy and reliability, the models struggle on suppressing either Overconfidence errors (hallucinations) or Abstention errors (wrong but safe), where the low Overconfidence rate seems more closely related to the high LCM scores, besides the high Confidence rate. This observation indicates that a high LCM score might imply a low risk of hallucination and high confidence of reliability. 

\section{Predicting the performance on test set without gt annotation}

Since LCM can be computed without gt annotation, it could used to predict MLLM's performance on test set of Benchmark if the corresponding gt annotations are not release. We test three top MLLMs from their families on the MMMU test set, which has over 9000 test samples compared around 900 samples of val set. The three tested MLLMs are Gemma3.0, InternVL-3.5-8B and Qwen3.0-VL-8B. The LCM scores on MMMU test set are presented in Table~\ref{tab:MMMU-val-test-comp}, compared with the LCM scores on val set. In each column, LCMval/LCMtest scores are presented for comparison. On the table, one can find that the LCM scores on test set are reasonably close to those on val set, and the ranking orders maintained. These results indicate that LCM could be employed to predict the performance and ranking on MLLMs on test set of a leaderboard Benchmark even though the gt annotations are not released.
\begin{table}
  \caption{Comparison of LCM scores on val and test sets of MMMU.}
  \label{tab:MMMU-val-test-comp}
  \footnotesize
  \centering
  \begin{tabular}{l|ccccccc}
    \toprule
    MLLM & Overall & Art \& & Business & Science & Health \& & Human \& & Tech \&  \\
     &  & Design &  &  & Medicine & Social Sci. & Eng.\\
    \midrule
    Gemma3.0-12B & .1216/.1083 & .2000/.2136 & .0746/.0849 & .1168/.0775 & .1241/.1379 & .2185/.2223 & .0469/.0376\\
    InternVL-3.5-8B & .3404/.3761 & .2819/.2905 & .3707/.4024 & .3511/.3902 & .3695/.3854 & .2980/.3450 & .3527/.3948 \\
    Qwen3.0-VL-8B & .4013/.4986 & .6156/.4511 & .2277/.5780 & .3570/.4683 & .4130/.5065 & .5621/.4849 & .3116/.5076 \\
    \bottomrule
  \end{tabular}
\end{table}
It may also be used to predict the performance of test set of public benchmark even though the gt annotation is not released.

\section{Time cost to obtain LCM}

To obtain LCM score, we introduce a set of YN tests on one MC sample, and a set of MC tests for one test unit in NaturalBench. Naturally, the time cost is increased. The average time cost for each test sample on NegBench, MMMU and NaturalBench are presented in Table~\ref{tab:time-cost}, where, for each Benchmark, the left column is the time for default test of the benchmark, and the right column is the time for additional tests for obtaining LCM.

\begin{table}
  \caption{Average time cost (seconds) for one sample on NegBench, MMMU, and NaturalBench.}
  \label{tab:time-cost}
  \footnotesize
  \centering
  \begin{tabular}{l|cc|cc|cc|cc}
    \toprule
      & \multicolumn{2}{c|}{NegBench-COCO} & \multicolumn{2}{c|}{NegBench-VOC} & \multicolumn{2}{c|}{MMMU} & \multicolumn{2}{c}{NaturalBench} \\
    \cmidrule(r){2-9}
    MLLM & MC & YNs & MC & YNs & MC & YNs & YNs & MCs \\
    \midrule
    Gemma3.0-12B & 0.42 & 7.52 & 0.42 & 7.20 & 3.44 & 216.63 & 4.17 & 10.20 \\
    InternVL-2.0-8B & 0.53 & 3.84 & 0.52 & 3.53 & 1.05 & 25.56 & 3.23 & 12.18 \\
    InternVL-2.5-8B & 0.55 & 3.82 & 0.53 & 3.50 & 0.84 & 14.39 & 3.21 & 9.36 \\
    InternVL-3.0-8B & 0.39 & 4.55 & 0.71 & 4.29 & 0.80 & 10.66 & 2.82 & 9.37 \\
    InternVL-3.5-8B & 1.71 & 5.04 & 1.09 & 4.62 & 1.32 & 32.32 & 3.46 & 9.77 \\
    LLaVA-1.5-13B & 0.44 & 1.80 & 0.44 & 1.75 & 5.10 & 21.04 & 2.64 & 4.78 \\
    LLaVA-1.6-13B & 0.57 & 2.83 & 0.51 & 2.86 & 7.27 & 22.11 & 1.79 & 9.31 \\
    Qwen-VL-7B-Chat & 0.59 & 1.39 & 0.56 & 1.30 & 0.53 & 1.76 & 1.31 & 4.79 \\
    Qwen2.0-VL-7B & 0.27 & 1.61 & 0.40 & 2.10 & 0.34 & 1.79 & 1.86 & 5.79 \\
    Qwen2.5-VL-7B & 0.32 & 2.95 & 0.26 & 2.45 & 0.35 & 1.76 & 5.36 & 11.12 \\
    Qwen3.0-VL-8B & 0.22 & 1.68 & 0.18 & 1.43 & 0.48 & 4.59 & 4.08 & 8.15 \\
    \bottomrule
  \end{tabular}
\end{table}

\section{Examples for visual examination}

Guided by low VL-LCM scores, user can focus on uncertain examples of the VQA cases on the same visual-language understanding and knowledge point for manual validation and justification even without gt annotation. 

\begin{figure}[t]
  \centering
\includegraphics[width=1.0\textwidth]{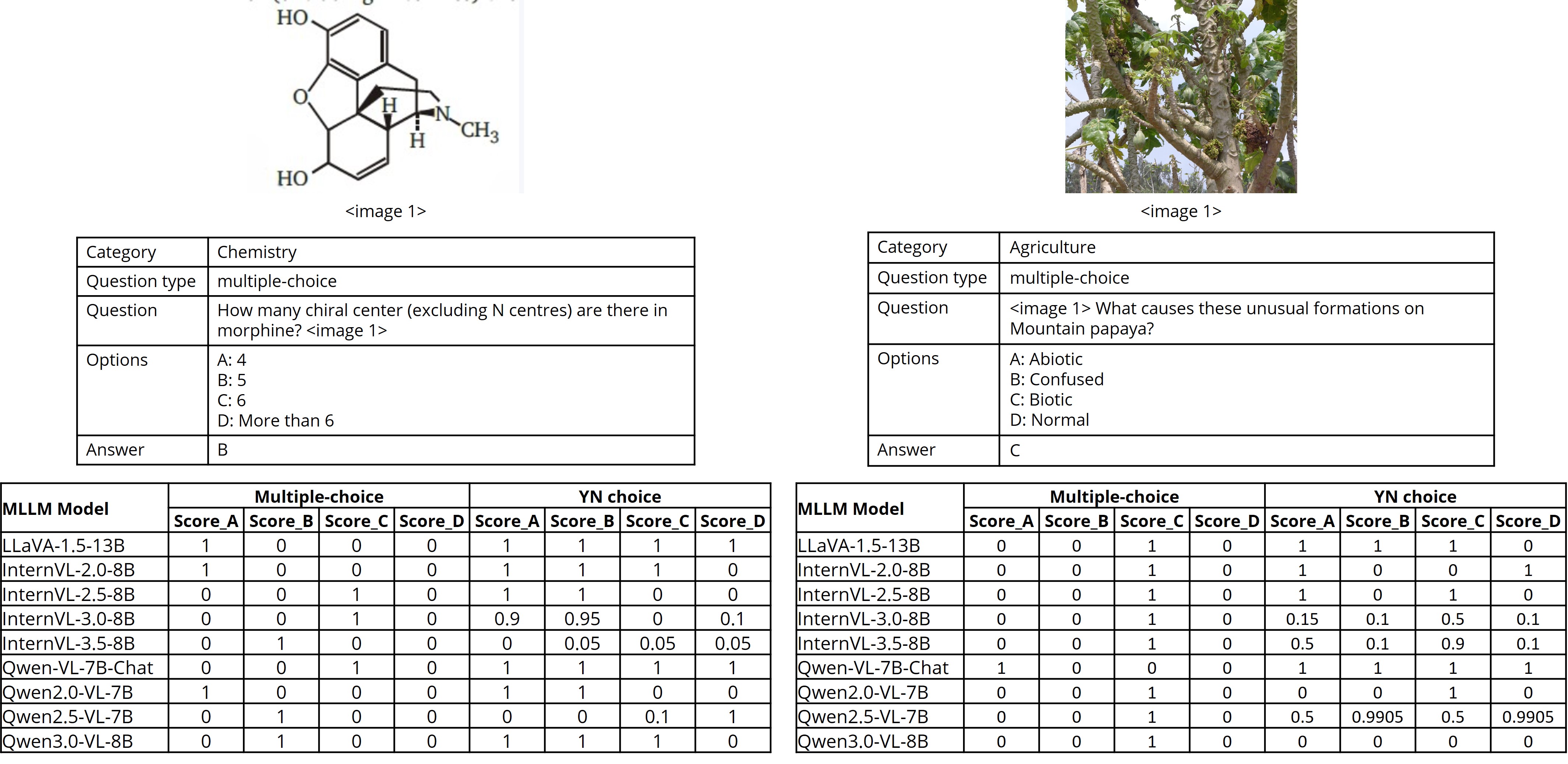} 
\caption{Two examples of low VL-LCM scores from MMMU, where Score means the probability of the prediction by the corresponding model. Many models are able to predict the correct answers in MC-VQA tests with default prompt format. However, when tested one-by-one on the YN-VQA questions, they produce logical inconsistent answers as observed in the table.}
\label{fig:mmmu_results}
\end{figure}

Two examples with low VL-LCM scores from MMMU are presented in~\ref{fig:mmmu_results}. On these two examples, when tested on MC-VQA format, one model might predict the correct answer exploiting the cues to select only one answer from multiple choices, as InternVL-3.5, Qwen2.5-VL, and Qwen3.0-VL models on the left sample, and most of the models on the right samples except Qwen-VL-7B-Chat. However, when tested on YN-VQA with positive and negative answers, or under sufficient and necessary cause-effect relations which require correct visual-language understanding of the problem, the models would produce inconsistent predictions, as shown in the responses in YN tests in the tables, leading to low VL-LCM scores.

\begin{figure}[h]
  \centering
\includegraphics[width=1.0\textwidth]{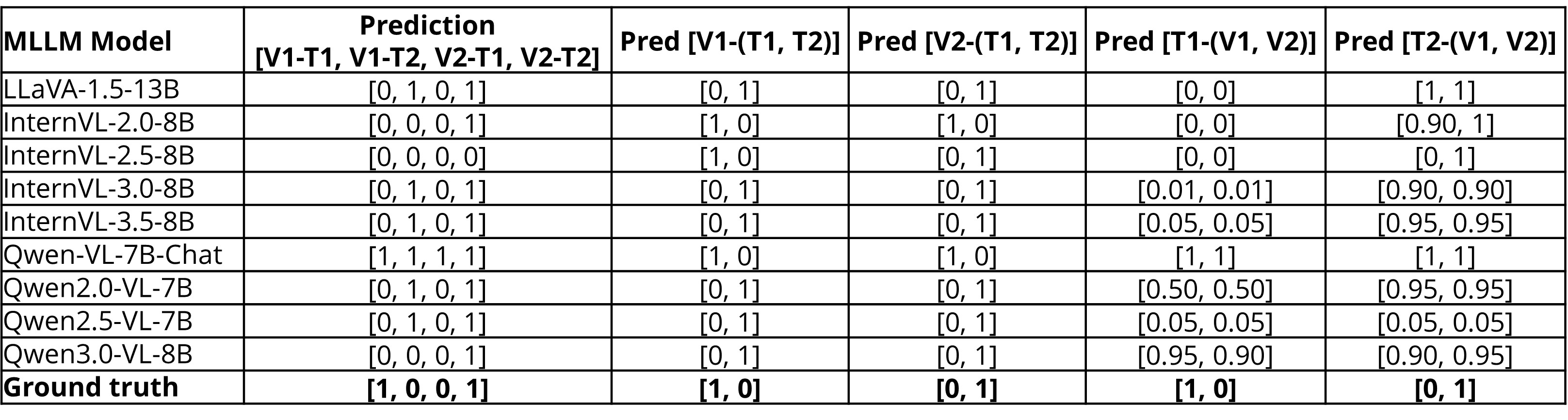} 
\caption{The prediction results of tested MLLMs on the samples from NatConBench shown in~\ref{fig:natconbench_mc}.}
\label{fig:results_mc}
  \centering
\includegraphics[width=1.0\textwidth]{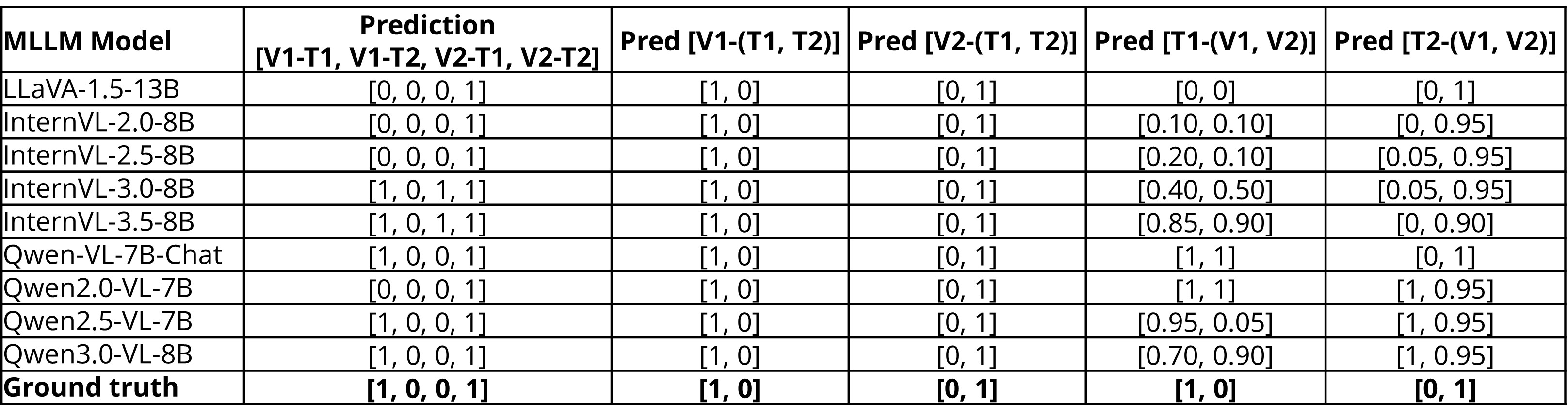} 
\caption{The prediction results of tested MLLMs on the samples from NatConBench shown in~\ref{fig:natconbench_tf}.}
\label{fig:results_tf}
\end{figure}

The evaluations on the two auto-generated samples of NatConBench are presented in~\ref{fig:results_mc} and~\ref{fig:results_tf}. One can observe the inconsistent predictions in both YN group tests and MC group tests. From the results, one can observe that the evaluated models may predict all correct answers when tested on four image-question pairs, but it is still very difficult for them to produce logically consistent predicts on both YN and MC tests, or under both sufficient and necessary conditions. The capability to produce answers with logic consistency on both sufficient and necessary conditions might be beyond the capability to simply predict a correct answer with high accuracy.

\end{document}